\documentclass[10pt,journal,compsoc]{IEEEtran}

\usepackage[switch]{lineno}
\usepackage{times}
\usepackage{epsfig}
\usepackage{graphicx}
\usepackage{amsmath}
\usepackage{amssymb}
\usepackage{algorithm}
\usepackage{algpseudocode}
\usepackage{float}
\usepackage{color}
\usepackage{colortbl}
\usepackage{breqn}
\usepackage{rotating}
\usepackage{times}
\usepackage{epsfig}
\usepackage{enumitem}
\usepackage{makecell}
\usepackage{lineno}
\usepackage{tabularx}
\usepackage{multirow}
\usepackage{graphicx}
\usepackage{hhline}
\usepackage{amsmath}
\usepackage{amssymb}
\usepackage{amsfonts}
\usepackage{empheq}
\usepackage{float}
\usepackage{framed, color}
\usepackage{graphics}
\usepackage{graphicx}
\usepackage[]{graphicx}
\usepackage{epsfig} 
\usepackage{color}
\usepackage{filecontents}
\usepackage{multirow}
\usepackage{filecontents}
\usepackage{verbatim}
\usepackage{tabularx}
\usepackage{lineno}
\usepackage{setspace}
\usepackage{textcomp,booktabs}
\usepackage{caption}
\usepackage{stmaryrd}
\usepackage[mathscr]{eucal}
\usepackage{lineno}
\usepackage{subfigure}
\usepackage{subcaption}
\usepackage[export]{adjustbox}
\usepackage{xurl}

\usepackage{times}
\usepackage{epsfig}
\usepackage{graphicx}
\usepackage{amsmath}
\usepackage{amssymb}
\usepackage{xr-hyper}
\usepackage{hyperref}
\usepackage[capitalize]{cleveref}
\usepackage[dvipsnames]{xcolor}

\def\0{{\bf 0}}
\def\1{{\bf 1}}

\definecolor{hyperlinkTextType}{rgb}{0.8, 0.0, 0.0}

\definecolor{red}{rgb}{0.95,0.4,0.4}
\definecolor{purered}{rgb}{1,0,0}
\definecolor{blue}{rgb}{0.4,0.4,0.95}
\definecolor{darkblue}{rgb}{0,0,0.8}
\definecolor{darkred}{rgb}{1,0,0}
\definecolor{darkgreen}{rgb}{0,0.5,0}
\definecolor{grey}{rgb}{0.6,0.6,0.6}
\definecolor{col1}{RGB}{232, 161, 148}
\definecolor{col2}{RGB}{148, 187, 232}
\definecolor{lightgrey}{rgb}{0.85,0.85,0.85}
\definecolor{lightlightgrey}{rgb}{0.9,0.9,0.9}
\definecolor{verylightBG}{rgb}{0.9,0.99,0.99}
\definecolor{darkgreen}{rgb}{0.3, 0.75, 0.3}

\definecolor{col33}{RGB}{206, 239, 255}
\definecolor{lightgray}{rgb}{0.85,0.85,0.85}
\definecolor{lightlightgray}{rgb}{0.9,0.9,0.9}
\definecolor{verylightBG}{rgb}{0.9,0.99,0.99}
\definecolor{darkgreen}{rgb}{0., 0.7, 0.2}

\graphicspath{{../figures/}}

\ifCLASSOPTIONcompsoc
  \usepackage[nocompress]{cite}
\else
  \usepackage{cite}
\fi

\ifCLASSINFOpdf
\else
\fi

\begin{document}

\title{Visual Species Recognition with Large Multimodal Models as Post-Hoc Correctors}

\author{
Tian Liu$^{*}$,
Anwesha Basu$^{*}$,
James Caverlee,
Shu Kong$^{\dag}$
\IEEEcompsocitemizethanks{\IEEEcompsocthanksitem 
$^*$Equal Contributions. $^{\dag}$Corresponding author.
\IEEEcompsocthanksitem 
Tian Liu, Anwesha Basu, and James Caverlee are with the Department of Computer Science and Engineering, Texas A\&M University.
E-mail: \{ltmask, wesha.basu, caverlee\}@tamu.edu
\IEEEcompsocthanksitem 
Shu Kong is with the Department of AI, University of Macau.
E-mail: skong@um.edu.mo
}
}

%



\IEEEtitleabstractindextext{%
\begin{abstract}
Visual Species Recognition (VSR) is a fundamental task in scientific disciplines that require species-level identification, including ecology, palynology, evolutionary biology, systematics, and phylogenetics.
Automating VSR through machine learning can significantly accelerate these efforts. 
However, species-level annotation requires extensive domain expertise, 
making large-scale labeled datasets difficult to obtain. 
Consequently, few-shot learning (FSL) is a practical paradigm, 
where an expert model is trained using only a few labeled examples.
Meanwhile, Large Multimodal Models (LMMs) have demonstrated unprecedented zero-shot visual recognition capabilities,
raising the question of whether they can serve as an alternative to FSL expert models for VSR.
We start this work with a systematic comparison between FSL expert models and LMMs,
revealing that,
despite advanced prompting strategies, 
contemporary LMMs significantly underperform FSL expert models.
Interestingly, we find that LMMs possess a complementary strength:
given an image and a shortlist of candidate species generated by an expert model,
LMMs can often recover the correct label when the expert model's top prediction is incorrect.
Motivated by this,  
we propose Post-hoc Correction (POC), 
a simple training-free framework that leverages an LMM to post-process an expert model's top predictions.
We develop a multimodal prompting strategy to enable POC to improve FSL expert models
by 6.4 accuracy points, averaged over five VSR benchmarks.
We show that POC generalizes across diverse FSL methods, visual encoders, and LMMs,
making it a practical and effective framework for VSR.

\end{abstract}

}

\maketitle

\IEEEdisplaynontitleabstractindextext

%
\IEEEpeerreviewmaketitle

\IEEEraisesectionheading{
\section{Introduction}
\label{sec:intro}
}

\IEEEPARstart{V}isual Species Recognition (VSR), the task of identifying species of plants or animals from imagery,
is fundamental to various disciplines such as  
evolutionary biology~\cite{mayr1970populations, romero2020improving},
ecology~\cite{borowiec2022deep, lurig2021computer}, 
palynology~\cite{germeraad1968palynology, Adaime2025pollen},
systematics~\cite{mayr1999systematics}, 
and phylogenetics~\cite{nixon1990amplification, barraclough2001phylogenetics, zuntini2024phylogenomics, adaime2024deep}.
VSR relies on extensive domain expertise and is routinely performed manually by specialists \cite{hopkins2002declines}.
Using machine learning to automate VSR can substantially reduce this burden and accelerate research across these fields.

\begin{figure*}[t]
  \centering
  \includegraphics[width=0.995\linewidth, clip=true,trim = 0mm 0mm 0mm 0mm]{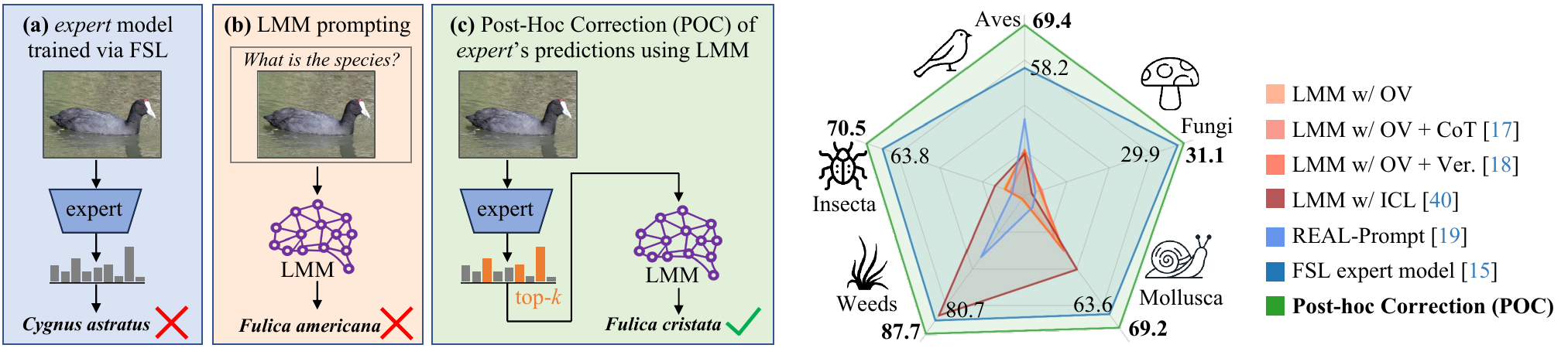}
  \vspace{-2mm}
  \caption{
  \small
  \textbf{Summary of methods and results.}  
  We compare two types of VSR methods: 
  \textbf{(a)} 
  an FSL expert model by finetuning a VLM's (CLIP \cite{radford2021learning, cherti2023reproducible}) visual encoder on 16-shot labeled imagery \cite{liu2025few},
  and \textbf{(b)} 
  an LMM (Qwen-2.5-VL-7B-Instruct \cite{qwen2.5-vl}) instructed with different prompting strategies \cite{kojima2022large, weng2023large}.
  Somewhat surprisingly, the LMM significantly underperforms the FSL expert model (on the right panel).
  However, we show that (1) although the expert may fail to predict the correct label as its top-1 choice, the correct label is often within the top-$k$ predictions (\cref{fig:handpicked_examples}), 
  and (2) LMMs perform substantially better when asked to select the most probable species from a small set of candidate species than from an large or unconstrained label space (\cref{tab:lmm_failue}).
  \textbf{(c)} 
  Motivated by these observations,
  we propose \textbf{Post-hoc Correction (POC)}, a simple framework that leverages an LMM to refine the expert models' predictions (\cref{fig:poc_workflow}) in a post-hoc manner. 
  The right panel shows that, across five distinct benchmarks,
  POC significantly outperforms the compared FSL expert models and LMM-based methods.
  }
  \vspace{-1mm}
  \label{fig:teaser}
\end{figure*}

Training machine-learning models typically requires large-scale labeled datasets.
For VSR, however,
acquiring species-level annotations is costly and requires substantial domain expertise.
Moreover, 
many species are rare and difficult to capture in sufficient numbers \cite{parashar2024neglected, su2021realistic}.
Consequently, it is realistic for domain experts to label a small number of images,
motivating the use of few-shot learning (FSL) methods to train ``\textit{expert}'' models for VSR.
Contemporary FSL methods adapt pretrained backbones using limited labeled data~\cite{liu2025few, clap24, wang2025robust, liu2025solving}.
These backbones can be either the visual encoder of a Vision-Language Model (VLM) \cite{radford2021learning} or a Vision Foundation Model (VFM)~\cite{oquab2023dinov2}.
Representative adaptation strategies include
prompt learning \cite{maple, zhou2022learning, yao2023visual},
adapter learning \cite{taskres, clipadapter, tipadapter}, 
linear probing~\cite{clap24, lin2023multimodality}, 
and finetuning \cite{liu2025few}.
Notably, simple finetuning of a pretrained backbone has emerged as a surprisingly strong FSL method (\cref{fig:teaser}a), often outperforming more sophisticated ones~\cite{liu2025few}.

To automate VSR,
an alternative to FSL expert models is to leverage Large Multimodal Models (LMMs) \cite{maruf2024vlm4biobenchmarkdatasetevaluate, stevens2025minddatagapevaluating}, 
which underpin modern AI systems \cite{achiam2023gpt, qwen, vteam2025glm}.
LMMs have become increasingly accessible and have demonstrated unprecedented zero-shot visual recognition capabilities.
LMMs are often extremely large and proprietary, many with access restricted to APIs,
making finetuning LMMs on downstream data impractical.
Yet, 
users can tune prompts~\cite{menon2022visual, parashar2023prompting} to adapt LMMs to specific visual recognition tasks.
The recent work \cite{maruf2024vlm4biobenchmarkdatasetevaluate}
examines LMMs for VSR in settings of zero-shot recognition and multiple-choice questions.
However, 
it is unclear how LMMs perform when compared with FSL expert models.
We start our work by addressing this question through experiments.
We summarize key findings below.

We evaluate the recent LMM, Qwen-2.5-VL-7B-Instruct \cite{qwen2.5},
on five VSR benchmarks (detailed in \S\ref{sec:problem}).
For each test image,
we prompt the LMM with the open-vocabulary question
{\em ``What species is shown in the image?''} (\cref{fig:teaser}b).
Aiming for more accurate inference,
we further instruct the LMM to explicitly output its reasoning process,
following prior work showing the effectiveness of such prompting strategies~\cite{kojima2022large, weng2023large}.
We also design an in-context learning prompt~\cite{jiang2405many, agarwal2024many} that constrains the prediction space to all candidate species in the benchmarks
by providing both scientific and common names as the  context~\cite{parashar2023prompting}.
Interestingly, these strategies yield limited improvements (ref.  \cref{fig:teaser}-right). 
Overall, the LMM significantly underperforms
the simple FSL expert model \cite{liu2025few} (\cref{fig:teaser}a),
which finetunes the visual encoder of the VLM OpenCLIP ViT-B/32 \cite{cherti2023reproducible} using only a small number of labeled examples (16 shots per class; see \S\ref{ssec:FSL}).

A deeper analysis of the LMM evaluation results reveals two key insights.
First, although an FSL expert model may fail to predict the correct species as its top-1 choice, the correct species is often included among its top-$k$ predictions (e.g., $k=5$).
Second, 
LMMs perform substantially better when instructed to select the most probable species from a small set of candidate species rather than from a large or unconstrained label space.
Building on these two insights, we propose leveraging LMMs to post-hoc correct predictions of the FSL expert models.
Specifically, 
given a test image,
we first obtain the top-$k$ predictions of an FSL expert model.
We then construct a prompt to ask LMM to select the most probable one from these candidate species as the final prediction.
Furthermore, 
we investigate a variety of prompting strategies and develop a simple yet effective multimodal prompt that instructs the LMM to re-rank the top-$k$ predictions of the expert model, by additionally providing their confidence scores and corresponding few-shot training images.
We name our method \textbf{Po}st-hoc \textbf{C}orrection (\textbf{POC}).
As demonstrated in our experiments (summarized in \cref{fig:teaser}-right), POC consistently improves the FSL expert models.
Notably, POC is training-free, validation-free, model-agnostic, and free of human intervention.

{\bf Contributions}.
We make three key contributions:
\begin{enumerate}[topsep=0pt, partopsep=0pt, leftmargin=0.3in]
\item 
    We establish a unified protocol that enables fair evaluation of FSL expert models and LMMs for VSR.
    Our experiments reveal that,
    despite the significant advancement of LMMs, FSL expert models remain substantially more accurate.

\item 
    Building on the insight that LMMs can effectively correct FSL expert's predictions,
    we propose Post-hoc Correction (POC),
    a simple framework that employs an LMM to effectively refine expert models' predictions.
    
\item 
    We show that POC can be readily integrated with diverse FSL methods across different backbones of expert models and LMMs,
    highlighting that it is a practical framework for solving VSR.

\end{enumerate}

\section{Related Works}
\label{sec:related}

\begin{figure}[!t]
\centering
\includegraphics[width=\linewidth]{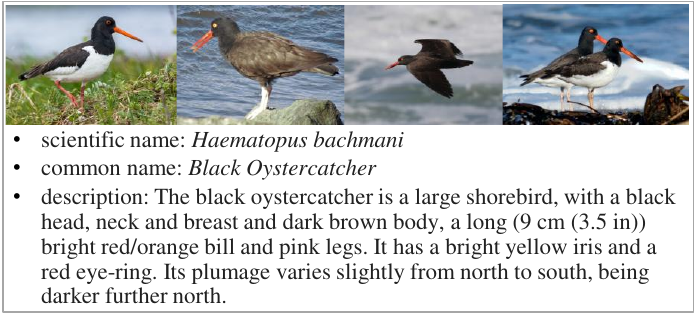}
\vspace{-6mm}
\caption{
\small
{\bf An illustrative snippet of expert-crafted annotation guidelines.}
In real-world annotation practices,
domain experts craft an instruction to specify each predefined species
with scientific and common names,
descriptions of characteristic traits,
and a few representative images.
These multimodal data can serve as a few-shot training set to develop VSR models. 
}
\label{fig:annotaion-example}
\vspace{-4mm}
\end{figure}

{\bf Visual Species Recognition (VSR)} is a fundamental task across numerous scientific disciplines, including  
evolutionary biology~\cite{mayr1970populations},
ecology~\cite{schluter2001ecology}, 
palynology~\cite{germeraad1968palynology},
systematics~\cite{mayr1999systematics}, 
and phylogenetics~\cite{nixon1990amplification}.
It requires domain expertise to identify organisms from visual observations at the species level.
Prior research has predominantly studied VSR through fine-grained visual categorization (FGVC)~\cite{duan2012discovering, zhang2014part, lin2015bilinear, kong2017low},
where models are trained via supervised learning on massive labeled images.
However, 
obtaining species-level annotations at scale is prohibitively expensive and, for rare species, often infeasible due to scarce imagery \cite{su2021realistic, parashar2024neglected}.
In practice, 
it is more realistic for experts to specify species through a few exemplar images accompanied by textual descriptions (\cref{fig:annotaion-example}).
This naturally casts VSR as a multimodal few-shot learning (FSL) problem \cite{madan2024revisiting, liu2025few, picek2025fungitastic},
which forms the setting of our work.
Concurrently, several recent studies have developed foundation models (FMs) tailored for VSR \cite{stevens2024bioclip, gu2025bioclip2, zhang2026biocapexploitingsyntheticcaptions, sastry2024taxabindunifiedembeddingspace}.
These FMs are typically trained on tens of millions of labeled bio-images sourced from existing datasets annotated by experts and citizen scientists,
reflecting a significant prior annotation effort.
In contrast,
we study VSR in the multimodal FSL setting,
benchmarking both expert models trained via FSL and publicly available general-purpose Large Multimodal Models (LMMs), without relying on meticulously curated data to train VSR-specific FMs.

{\bf Few-Shot Learning (FSL).}
Contemporary FSL methods largely build upon pretrained Vision-Language Models (VLMs).
Most approaches keep the VLM frozen  and learn only a small number of additional parameters, such as  
prompt tokens~\cite{zhou2022learning, zhou2022conditional, yao2023visual, wang2025attention}
or lightweight adapters~\cite{gao2023clip, zhang2022tip, song2023meta, zhang2024dual}.
More recently, several studies \cite{wortsman2022robust, goyal2023finetune, kumar2022finetuning, liu2025few, liu2025solving, wang2025robust}
have demonstrated that directly finetuning VLMs on few-shot training images achieves superior performance, without suffering from severe overfitting, compared with prompt-tuning and adapter-based alternatives.
Notably, 
recent FSL research \cite{madan2024revisiting, liu2025few} has advocated a more realistic evaluation protocol inspired by practical data annotation workflows,
where domain experts specify object categories in annotation guidelines for non-expert annotators to follow.
Such guidelines typically contain a small set of visual examples accompanied by textual descriptions (\cref{fig:annotaion-example}),
naturally leading to a multimodal FSL setting.
This protocol closely matches the VSR scenario,
in which domain experts characterize each species using a few representative images and trait descriptions.
Accordingly, 
we experiment with state-of-the-art FSL methods that finetune VLMs using both few-shot images and texts.
Furthermore,
we leverage the same multimodal information to instruct modern LMMs for VSR and compare their performance with the FSL model.
These investigations yield key insights that motivate the development of our approach.

{\bf Foundation Models} (FMs), 
pretrained on web-scale data,
have significantly advanced numerous downstream vision tasks.
Among them,
Vision-Language Models (VLMs) learn a shared feature embedding space for images and text from massive image-caption corpora \cite{radford2021learning, align, cherti2023reproducible},
enabling zero-shot and open-vocabulary visual recognition \cite{sun2024alpha}.
A VLM consists of a visual encoder and a text encoder and has become a backbone in various downstream recognition tasks.
Vision Foundation Models (VFMs), such as DINOv2 \cite{oquab2023dinov2}, are pretrained through self-supervised learning on images \cite{caron2021dino, hinton2015distilling} and capture rich fine-grained visual features.
Both VLMs and VFMs have moderate parameter sizes and 
demonstrated strong transferability across downstream tasks \cite{clap24, lin2023multimodality, zhang2025revisiting, mai2025lessons, zhang2025finer}.
They can be effectively finetuned on the few-shot data to the expert model for VSR \cite{liu2025few}.
In parallel, 
Large Multimodal Models (LMMs) integrate large language models (LLMs) with modality-specific encoders,
resulting in a substantially larger model size with powerful multimodal reasoning capabilities.
Examples include open-source models such as LLaVA \cite{liu2023visual}, InternVL \cite{chen2024internvl}, and Qwen2.5-VL \cite{qwen2.5-vl},
as well as proprietary systems such as 
Flamingo \cite{alayrac2022flamingo}, Gemini \cite{team2023gemini}, and GPT-4V \cite{achiam2023gpt}.
LMMs excel at complex tasks, such as Visual Question Answering (VQA) \cite{antol2015vqa, liu2023visual, NIPS2017_f61d6947} and image captioning \cite{you2016image, sharma2018conceptual}.
However, their large model size and, in many cases, restricted API-only access make task-specific finetuning impractical.
Recent studies have evaluated LMMs for VSR 
in either zero-shot settings \cite{stevens2025minddatagapevaluating} or by comparing them with classifiers trained on features from frozen VLMs \cite{maruf2024vlm4biobenchmarkdatasetevaluate}.
Nevertheless,
existing work has neither explored multimodal few-shot data to instruct LMMs for VSR, nor leveraged/compared state-of-the-art FSL methods.
In this work,
we bridge this gap by comparing both paradigms.
Through comprehensive empirical evaluation,
we identify their respective strengths and integrate them with a simple yet effective approach.

{\bf Visual Recognition using LMMs.}
We review recent methods that leverage LMMs for image recognition.
Beyond directly prompting LMMs with queries such as ``What species is shown in this image?'', 
prior works exploit LMMs to generate auxiliary semantic information, including visual attributes \cite{menon2022visual, CUPL, yan2023learning, saha2024improved, maniparambil2023enhancing},
species synonyms \cite{parashar2023prompting, parashar2024neglected},
descriptive captions \cite{liu2024democratizing, yang2022empirical}, 
and structured scene graphs \cite{MitraCCoT}, 
which are subsequently used to improve recognition performance.
More recently, 
researchers have sought to enhance visual recognition by exploiting the reasoning capabilities of LMMs \cite{wei2022chain, kojima2022large, wang2023self, liu2023visual}
through techniques such as 
Chain-of-Thought prompting \cite{lu2025rsvp, brown2020language, kojima2022large}, 
self-verification \cite{wang2023self}, 
and in-context learning \cite{agarwal2024many, jiang2405many, brown2020language, min2022rethinking, zhang2023makes}.
A concurrent study \cite{liu2024rar} employs an LMM to rerank candidate classes selected according to CLIP feature similarities.
While our method similarly employs an LMM to rerank the top predictions produced by an FSL expert model,
our key novelty lies in the \textit{multimodal in-context prompting} strategy we designed, which integrates the expert model’s predictions, confidence scores, the corresponding few-shot training images, and relevant species knowledge.
Our prompt yields significantly better performance.

\section{Problem Formulation}
\label{sec:problem}

We start with the problem formulation of VSR, followed by the benchmark datasets.

\begin{figure}[!t]
\centering
\includegraphics[width=\linewidth]{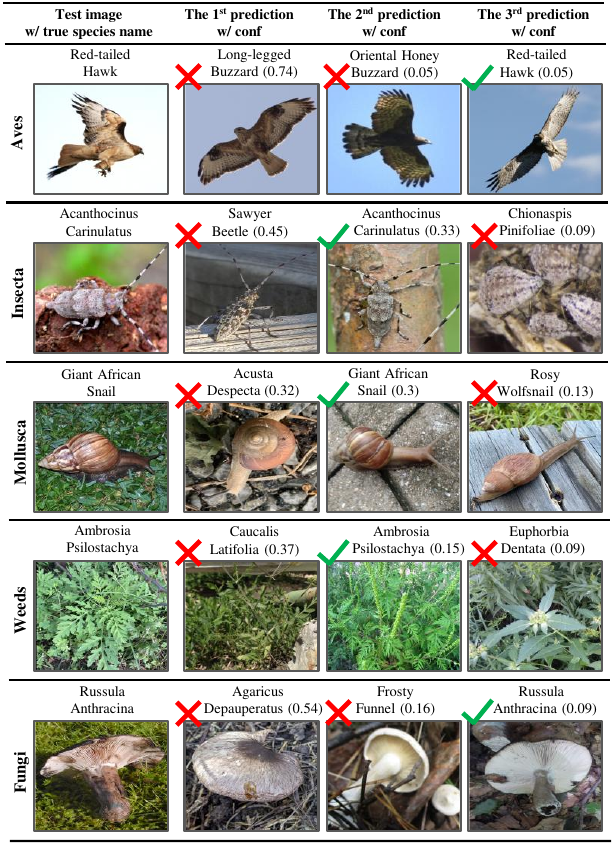}
\vspace{-7.5mm}
\caption{
\small
\textbf{Examples of test images from five VSR benchmarks,} 
along with an expert model's top-3 predicted species and the confidence scores.
A reference image is displayed for each predicted species.
We train an expert model by finetuning the visual encoder of the OpenCLIP ViT-B/32 model \cite{cherti2023reproducible} on 16-shot training data following \cite{liu2025few}.
The prevalence of visually similar species among the top-3 predictions underscores the significant challenges of the VSR task.
Notably, although the top-1 predictions are incorrect,
the top-3 predictions often contain the correct species.
}
\label{fig:handpicked_examples}
\vspace{-1mm}
\end{figure}

{\bf Problem Formulation.}
Visual species recognition (VSR) aims to classify an input image $I$ into one of $C$ predefined taxa, 
indexed by $y = \{1, 2, \ldots, C\}$. 
The taxa may represent fine-grained biological species that differ in subtle morphological attributes, such as beak curvature, wing patterns, or petal arrangements \cite{pinho2022, inat2021} (see \cref{fig:handpicked_examples}).
The formulation mirrors real-world annotation practices \cite{madan2024revisiting}, 
where domain experts specify each taxon using a few representative images and textual descriptions, including scientific and common names and characteristic traits (ref. Fig.~\ref{fig:annotaion-example}).
Consequently, 
such multimodal data constitute a few-shot learning training set, 
serving as the basis for developing VSR methods.
To evaluate each VSR method,
we report the averaged per-class accuracy on held-out test sets of benchmark datasets.

{\bf Benchmark Datasets.}
We set up benchmarks under the multimodal few-shot setting by 
sourcing data from existing fine-grained species classification datasets.
Specifically, from the Aves subset \cite{semi-aves} of iNaturalist 2018 (under the CC BY-NC license) \cite{van2018inaturalist},
we obtain the ``Aves'' benchmark, containing $m$ examples per bird species ($m = 4, 8, 16$) across 200 species.
From Species196 \cite{he2023species196} (under the CC BY-NC-SA 4.0 license), we construct three additional benchmarks: ``Insecta'', ``Weeds'', and ``Mollusca'', each consisting of
$m$ examples per species from the corresponding taxonomic groups, covering 78, 20, and 7 species, respectively.
From FungiTastic~\cite{picek2025fungitastic} (under CC BY 4.0 and CC BY-NC 4.0 licenses), we obtain the ``Fungi'' benchmark by sampling 
$m$ labeled examples per species, totaling 196 species. \cref{fig:handpicked_examples} shows representative images from these benchmarks.
For evaluation, we use the official test splits of each dataset. 
Evaluation is done on a held-out test split constructed from each dataset.
We report results using the 16-shot setting in the main paper and include results for the 4-shot and 8-shot settings in Supplement \cref{tab:improve_fsr_detail}, where conclusions are consistent.
Furthermore,
Supplement  \cref{tab:nonvsr_generalization} demonstrates that our approach generalizes to 
popular non-VSR datasets, including FGVC Aircraft \cite{aircraft}, Stanford Cars \cite{cars}, Describable Textures Dataset (DTD) \cite{dtd}, and Food-101 \cite{food}.

\section{Methodology}
\label{sec:methodology}

Given a few-shot training set, 
one can employ FSL methods to train an expert model.
Contemporary FSL approaches adapt foundation models, such as the VLM OpenCLIP \cite{cherti2023reproducible} or the VFM DINOv2  \cite{oquab2023dinov2}.
Notably, the recent method \cite{liu2025few} directly finetunes FMs and achieves superior performance compared with those that either learn prompts or lightweight adapters on top of a frozen pretrained backbone.
Alternatively,
one can leverage LMMs,
such as open-source models Qwen \cite{qwen2.5-vl} and InternVL \cite{wang2025internvl3_5},
and proprietary models like Gemini \cite{team2023gemini} and GPT~\cite{achiam2023gpt, openai2025gpt5} through APIs. 
Briefly, LMM-based approaches prompt an LMM to identify the species shown in a test image.
The prompt can be enriched 
with textual descriptions or other auxiliary information.
Next, we delve into these two types of methods and analyze their strengths and limitations.

In this section, we introduce the state-of-the-art FSL method (\S\ref{ssec:FSL}) and LMM-based prompting approaches  (\S\ref{ssec:LMM}), respectively,
followed by an analysis of their performance in \S\ref{ssec:insights} that uncovers actionable insights.

{
\setlength{\tabcolsep}{0.3em}
\begin{table}[t]
\centering
\small
\caption{\small
\small
{\bf VSR accuracy by different methods.}
As a reference, we report the performance of the zero-shot recognition method
ZS-VLM \cite{parashar2024neglected}, which uses the pretrained VLM OpenCLIP ViT-B/32.
Then, we report the few-shot learning method \cite{liu2025few},
which trains expert models for VSR in the 16-shot setting using the same VLM. 
We report LMM-based methods using the Qwen-2.5-VL-7B-Instruct, with open-vocabulary prompting (optionally augmented by
CoT or self-verification),
and In-Context Learning (ICL). 
Superscripts denote gains (\textcolor{Green}{green}) or degradations (\textcolor{Red}{red}) relative to open-vocabulary prompting. 
Refer to \S\ref{ssec:insights} for actionable insights.
}
\vspace{-3mm}
\label{tab:lmm_failue}
\scalebox{0.85}{
\begin{tabular}{lllllllll}
\toprule
 method  & Aves & Fungi & Insecta & Weeds & Mollusca &  \cellcolor{col33}mean\\
\#classes & (200) & (196) & (78) & (20) & (7) &  \cellcolor{col33}accuracy \\
\midrule

ZS-VLM \cite{parashar2024neglected} &44.8 &2.1 &10.0 &47.2 &19.7 &\cellcolor{col33}24.8 \\

FSL expert \cite{liu2025few} &58.2 &29.2 &63.8 &80.7 &63.6 &\cellcolor{col33}59.2 \\
\midrule
open vocab. 
& 36.8 & 3.0 & 13.5 & 17.0 & 35.9 & \cellcolor{col33}21.2 \\

\quad + CoT \cite{kojima2022large} 
& 34.2$^{\textcolor{Red}{-2.6}}$  
& 3.4$^{\textcolor{Green}{+0.4}}$  
& 10.9$^{\textcolor{Red}{-2.6}}$  
& 14.6$^{\textcolor{Red}{-2.4}}$  
& 32.0$^{\textcolor{Red}{-3.9}}$  
& \cellcolor{col33}19.0$^{\textcolor{Red}{-2.2}}$  \\

\quad + Ver. \cite{weng2023large}   
& 36.9$^{\textcolor{Green}{+0.1}}$  
& 3.1$^{\textcolor{Green}{+0.1}}$ 
& 13.0$^{\textcolor{Red}{-0.5}}$ 
& 16.6$^{\textcolor{Red}{-0.4}}$ 
& 37.5$^{\textcolor{Green}{+1.6}}$
& \cellcolor{col33}21.4$^{\textcolor{Green}{+0.2}}$  \\

ICL \cite{jiang2405many}
& 35.9$^{\textcolor{Red}{-0.9}}$  
& 1.4$^{\textcolor{Red}{-1.6}}$  
& 17.3$^{\textcolor{Green}{+3.8}}$  
& 78.1$^{\textcolor{Green}{+61.1}}$  
& 45.2$^{\textcolor{Green}{+9.3}}$  
& \cellcolor{col33}35.6$^{\textcolor{Green}{+14.3}}$  \\
\bottomrule
\end{tabular}
}
\vspace{-1mm}
\end{table}}

\subsection{Training VSR Models via Few-Shot Learning} 
\label{ssec:FSL}

Among numerous FSL approaches,
the recent method \cite{liu2025few} directly finetunes a foundation model (e.g., the visual encoder of VLM) using few-shot training images per class.
It consistently outperforms prior FSL approaches that typically freeze the pretrained backbone and employ parameter-efficient adaptation strategies, such as learning prompt tokens
\cite{zhou2022learning, zhou2022conditional, jia2022visual, maple, wang2025attention}, lightweight adapters \cite{clipadapter, tipadapter},
or classifiers \cite{clap24, lin2023multimodality}
Owing to its simplicity and strong empirical performance, 
we adopt it as the state-of-the-art FSL method to train expert models for VSR. 
We follow the training protocol of \cite{liu2025few} and use the open-source VLM OpenCLIP ViT-B/32 as the backbone.
Specifically,
we initialize a linear classifier using the text embeddings of the $C$ species,
where each species is represented by both its scientific and common names \cite{parashar2023prompting, parashar2024neglected}.
We then jointly finetune the visual encoder and classifier on the few-shot training set using cross-entropy loss.
In addition to this FSL method \cite{liu2025few},
our experiments (\cref{tab:improve_fsr}) further compare against a diverse set of FSL approaches.

\textbf{Performance Overview.}
We evaluate the FSL expert model and compare it with the recent zero-shot recognition method ZS-VLM \cite{parashar2024neglected}.
As shown in the first two rows of \Cref{tab:lmm_failue},
the FSL expert model significantly outperforms the zero-shot approach, 
achieving 59.2\% mean accuracy versus 24.8\%.
The significant performance gains highlight the value of adapting foundation models using even a small amount of labeled training data.
It is worth pointing out that ZS-VLM achieves 62.9\% accuracy \cite{cherti2023reproducible} on the widely used ImageNet benchmark, which contains 1,000 object categories \cite{russakovsky2015imagenet}.
In contrast, its performance drops to 24.8\% on our benchmarks,
underscoring the significant challenges of VSR, even for 
state-of-the-art foundation models.

\subsection{LMM-based Prompting for VSR}
\label{ssec:LMM}

LMMs can process both visual and textual inputs and perform complex multimodal reasoning tasks, such as visual question answering, making them a natural choice for VSR. 
However, fine-tuning LMMs is often impractical, 
as many LMMs are extremely large and proprietary, with access restricted to APIs (e.g., GPT \cite{openai2025gpt5} and Gemini \cite{team2023gemini}). 
Consequently, existing approaches primarily rely on  prompting~\cite{menon2022visual, parashar2023prompting} and in-context learning (ICL) \cite{brown2020language, jiang2405many, min2022rethinking, zhang2023makes}. 
In this work, we evaluate the recent open-source LMM Qwen-2.5-VL-7B-Instruct \cite{qwen2.5-vl}. 
We begin with an open-vocabulary prompting strategy by asking the model, for a given test image, \emph{``What species is shown in the image?''} \cite{maruf2024vlm4biobenchmarkdatasetevaluate}.
The model's prediction is considered correct if it matches either the scientific name or the common name of the ground-truth species. 
We further enhance this prompting strategy with Chain-of-Thought (CoT) reasoning \cite{kojima2022large} by adding the instruction \emph{``Let's think step by step''}, or with self-verification (denoted as ``Ver.'') \cite{weng2023large},
which asks the model to reason about its prediction.
Finally, we evaluate an ICL-based prompting strategy \cite{agarwal2024many, jiang2405many} by supplementing the prompt with the list of all class names of each dataset and asking the LMM to select the most probable species for the test image. 
The corresponding prompt templates are provided in Supplement \S\ref{sec:prompt_templates}.

\begin{figure}[t]
    \centering
    \includegraphics[width=\linewidth, clip=true, trim=0mm 0mm 0mm 0mm]{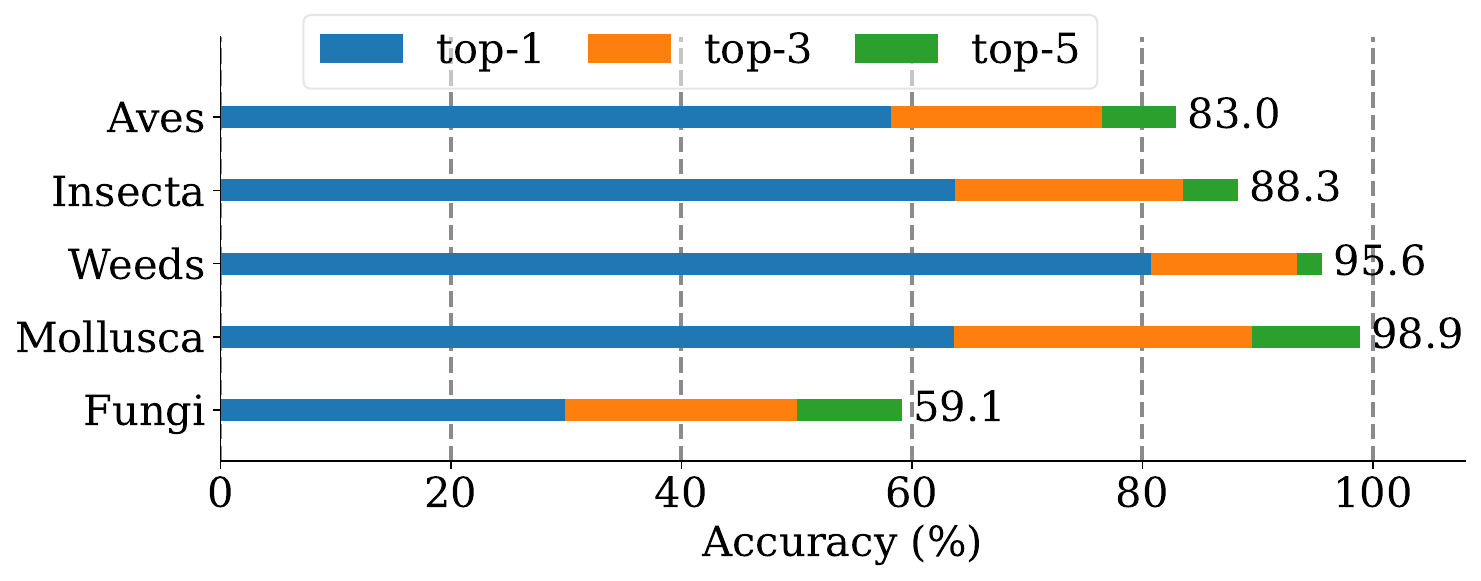}
    \vspace{-6.5mm}
    \caption{\small
    As is well known, a classifier's top-5 accuracy is typically much higher than the top-1 accuracy \cite{russakovsky2015imagenet}.
    That said, while an FSL expert model's top-1 prediction for a test image is incorrect,
    the ground-truth label is likely included in the top-$k$ predictions, e.g., $k=5$.
    This motivates our post-hoc correction method, which aims to recover the correct species from the expert model's top-$k$ predictions. 
    }
    \label{fig:topk_bar}
    \vspace{-1mm}
\end{figure}

\textbf{Performance Overview.}
\cref{tab:lmm_failue} compares the aforementioned LMM-based approaches with the FSL expert model. 
Somewhat surprisingly, 
all LMM-based methods substantially underperform the FSL expert model,
regardless of the prompting strategy employed.
Moreover, established reasoning techniques such as CoT \cite{kojima2022large} and self-verification \cite{weng2023large} yield marginal improvements over the naive open-vocabulary prompting and, in some cases, even degrade performance.
In contrast,
the In-Context Learning (ICL) prompting strategy \cite{jiang2405many}, which provides the complete set of candidate species names in the prompt, achieves more noticeable gains. These improvements are particularly pronounced on datasets with small species vocabularies, such as Weeds and Mollusca.
Nevertheless, even the strongest ICL-based results remain underperforming the FSL expert model, 
revealing a substantial performance gap between current LMM-based and FSL-based approaches for VSR.
Supplement \cref{tab:lmm_poc_summary} shows the same conclusions when using other LMMs such as GPT-5-Mini and GLM-4.1V-9B-Thinking.

\subsection{Actionable Insights}
\label{ssec:insights}

We next derive two key insights that motivate our approach. 
First, although the FSL expert model often produces incorrect top-1 predictions, its top-k predictions (e.g., even for small $k$=3) often include the ground-truth species (Figs.~\ref{fig:handpicked_examples} and \ref{fig:topk_bar}).
This observation implies that a suitable recovery mechanism could identify the correct species from the expert model's top-$k$ candidate set, 
thereby improving VSR performance.
Second, as shown in \cref{tab:lmm_failue},
ICL-based prompting outperforms open-vocabulary prompting on datasets that contain relatively fewer classes, such as Weeds and Mollusca.
The advantage of ICL diminishes as the number of classes increases (e.g., Aves and Fungi), likely because it becomes increasingly challenging to select the correct species from a large candidate vocabulary. 
This finding suggests that providing a small set of candidate species, rather than the entire label space, may substantially improve the effectiveness of LMMs for VSR.

Taken together, these two observations motivate the method introduced next, which combines the complementary strengths of an FSL expert model and the LMM-based ICL prompting. 
In brief, the expert model generates a small set of high-confidence candidate species, while the LMM reasons over these candidates to recover the correct prediction.

\subsection{The Proposed Method: Post-hoc Correction (POC)}
\label{ssec:POC}

\begin{figure}[!b]
\vspace{-4mm}
    \centering
    \includegraphics[width=1.0\linewidth]{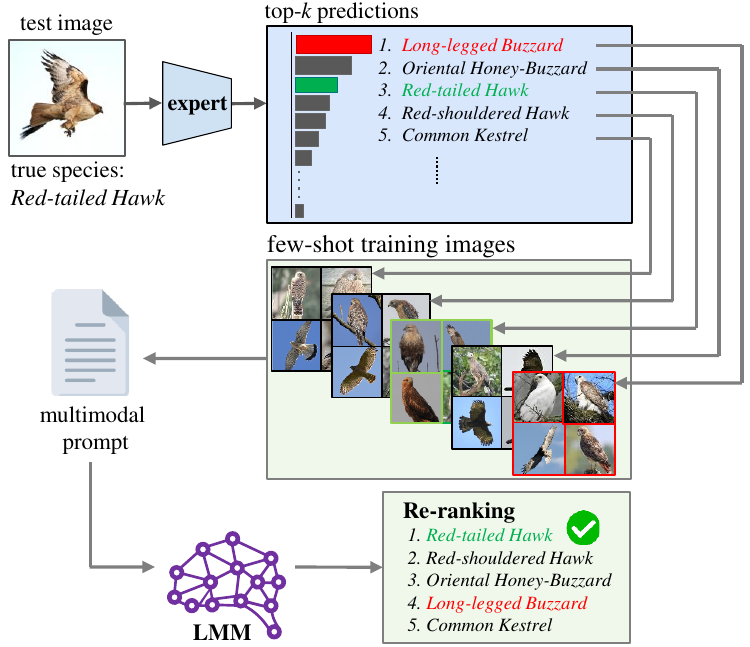}
    \vspace{-5.5mm}
    \caption{
    \small
    \textbf{The pipeline of Post-hoc Correction (POC).} 
    Given a test image, we run an expert model trained via FSL to obtain the top-$k$ predicted species and the corresponding confidence scores.
    Then, we construct a multimodal prompt using the test image, the scientific and common names of the $k$ species, their corresponding confidence scores, and the few-shot training images.
    We use the prompt to instruct an LMM to re-rank the candidate species and adopt the first one as the identified species for the test image.
    See the prompt template in \cref{fig:mmICL-prompt}.
    }
\vspace{-1mm}
\label{fig:poc_workflow}
\end{figure}

Building upon the aforementioned insights,
we develop a post-hoc correction method that uses an LMM to identify the correct species of a test image from the top-$k$ predictions of an FSL expert model.
We name our method \textbf{POC} and depict its pipeline in \cref{fig:poc_workflow}.
Below, we investigate a series of techniques to enhance its performance.

{
\setlength{\tabcolsep}{0.4em}
\begin{table}[t]
\centering
\small
\caption{\small
\small
{\bf Progressive development of the POC method.}
The expert model is trained by FSL \cite{liu2025few} with the backbone OpenCLIP ViT-B/32 \cite{cherti2023reproducible} in a 16-shot setting.
Using the LMM Qwen-2.5-VL-7B-Instruct \cite{qwen2.5-vl},
we progressively improve prompts to instruct the LMM to achieve better VSR,
refer to Supplement Table \ref{tab:abl_POC_v2} for details.
In sum, 
our final prompt 
contains 
the top-$k$ predicted species (combining both scientific and common names) and softmax confidence scores produced by the expert model, along with the corresponding few-shot training images.
It asks the LMM to re-rank the $k$ candidate species and choose the first one as the final species.
}
\vspace{-3mm}
\label{tab:ablation_mainpaper}
\scalebox{0.85}{
\begin{tabular}{lcccccl}
\toprule
 & Aves & Inse. & Weeds & Moll. & Fungi & \cellcolor{col33}mean acc. \\
\midrule

\rowcolor{gray!15}
FSL expert \cite{liu2025few} 
& 58.2 & 63.8 & 80.7 & \underline{63.6} & \underline{29.9} 
& \cellcolor{col33}59.2 \\

POC w/ top-5 species names 
& 63.8 & 40.1 & 81.3 & 46.9 & 16.2 
& \cellcolor{col33}49.7$^{\textcolor{Red}{-9.5}}$ \\

\quad + few-shot images 
& 68.9 & 61.1 & \textbf{88.6} & 56.4 & 27.7  
& \cellcolor{col33}60.5$^{\textcolor{Green}{+1.3}}$ \\

\quad \quad + confidence scores 
& \textbf{70.2} & \underline{67.4} & \underline{88.5} & 63.4 & 29.5   
& \cellcolor{col33}\underline{63.8}$^{\textcolor{Green}{+4.6}}$ \\

\quad \quad \quad + re-ranking 
& \underline{69.4} & \textbf{70.5} & 87.7 & \textbf{69.2} & \textbf{31.1}   
& \cellcolor{col33}\textbf{65.6}$^{\textcolor{Green}{+6.4}}$ \\

\bottomrule

\end{tabular}}
\vspace{-1mm}
\end{table}}

{\bf POC by Prompting with Top-$k$ Candidate Species.}
A straightforward baseline is to provide the LMM with the test image, along with the FSL expert model's top-$k$ predicted species, and ask it to select the most likely species from this candidate set.
The second row of Table~\ref{tab:ablation_mainpaper} reports this approach.
While it improves over the FSL expert model on the Aves and Weeds datasets, it substantially degrades performance on Insecta, Mollusca, and Fungi.
Overall, the mean accuracy drops from  59.2\% to 49.7\%,
falling well below that of the expert model.
The results suggest that simply prompting an LMM with the expert model's top-$k$ predictions is insufficient to effectively exploit the complementary strengths of the two models.
Nevertheless, the improvements observed on certain datasets indicate the potential of this direction,
motivating more technical innovations presented below.

\begin{figure*}[t]
  \centering
  \includegraphics[width=0.95\linewidth, clip=true, trim=0mm 0cm 0mm 0mm]{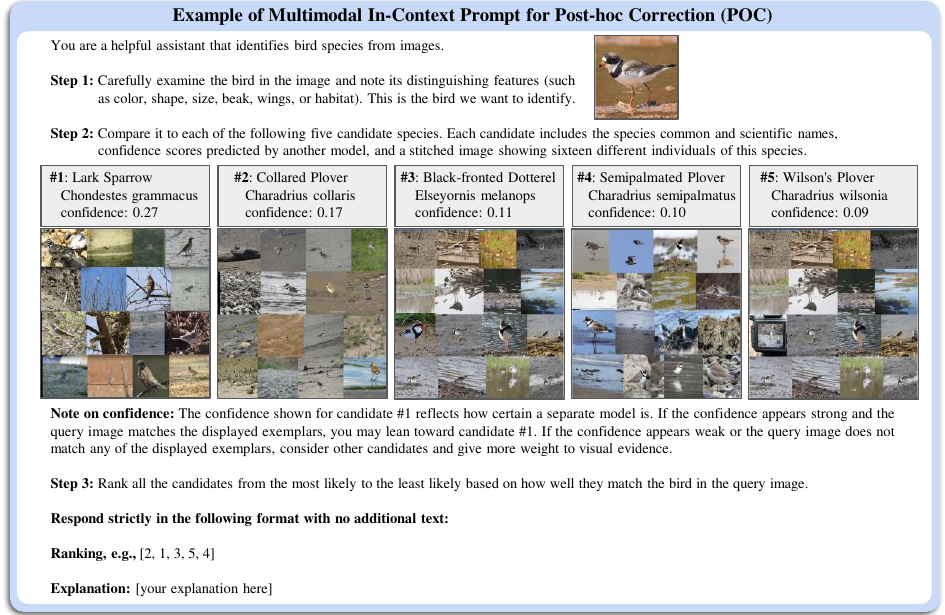}
  \vspace{-2mm}
  \caption{\small 
  \textbf{Illustration of the designed multimodal prompt in POC.}
  This prompt contains an expert model’s top-$k$ predicted species (e.g., $k=5$),
  along with their confidence scores and the few-shot training images (stitched in one image).
  The prompt asks the LMM to re-rank the candidate species by referring to the multimodal data and the test image.
  The top-1 re-ranked species is chosen as the final prediction.
  }
  \label{fig:mmICL-prompt}
  \vspace{-2mm}
\end{figure*}

{\bf POC by Incorporating Visual Examples into the Prompt.}
A natural extension of prompting with the expert's top-$k$ predicted species is to additionally provide their corresponding few-shot training images,
thereby constructing a multimodal prompt.
This strategy enables the LMM to compare the test image against visual references when selecting the most likely species from the candidate set. 
The third row of \cref{tab:ablation_mainpaper} demonstrates the effectiveness of this approach.
Compared with the text-only prompting baseline (second row),
incorporating visual examples yields significant accuracy gains,
confirming the value of visual evidence for species recognition.
Nevertheless,
the resulting performance only marginally surpasses that of the FSL expert model.
In particular, it underperforms the expert model on Insecta, Mollusca, and Fungi,
where the expert model's original top-1 predictions remain more reliable. 
These results suggest that further improvements can be achieved by more effectively exploiting expert-model outputs and LMM reasoning, as described next.

{\bf POC by Incorporating the Expert's Confidence Scores.}
Given the in-context reasoning capabilities of LMMs,
we hypothesize that POC can be further improved by enabling the model to adaptively weight the expert model's top-$k$ predictions according to their confidence scores.
To this end, we augment the multimodal prompt with the confidence scores associated with the expert model's top-$k$ predicted species. 
This addition allows the LMM to incorporate both visual evidence and the expert's uncertainty into its reasoning process.
As shown in the fourth row of \cref{tab:ablation_mainpaper}, adding confidence scores yields significant improvements across all datasets.

{\bf POC for Re-ranking vs. Selecting.} 
While the multimodal prompting strategy above already yields significant gains,
we further improve POC by asking the LMM to \textit{re-rank} the shortlisted candidate species instead of selecting the most likely one.
Re-ranking promotes a CoT-like reasoning process,
as the LMM must explicitly compare and order the candidates based on the provided context.
From the reranked candidates, we choose the first as the final identified species.
The last row of \cref{tab:ablation_mainpaper} shows that re-ranking generally improves over selection.
Our final prompt design, illustrated in \cref{fig:mmICL-prompt}, 
therefore instructs the LMM to re-rank candidate species based on the given test image,
the top-$k$ candidates' scientific and common names, 
the corresponding few-shot training images,
and the confidence scores produced by the FSL expert model.

\begin{table*}[t]
\centering
\small
\caption{\small
{\bf POC significantly improves existing FSL methods.}
We run POC with the LMM Qwen-2.5-VL-7B \cite{qwen2.5-vl}, and expert models learned by different FSL methods on the OpenCLIP ViT-B/32 backbone \cite{cherti2023reproducible} in the 16-shot setting.
These FSL methods include prompt learning \cite{zhou2022learning, maple}, adapter learning \cite{clipadapter, tipadapter}, linear probing  \cite{radford2021learning, lin2023multimodality, clap24}, and few-shot full finetuning  \cite{liu2025few}.
Additionally, FineR \cite{liu2024democratizing} is an LMM-based fine-grained recognition approach that integrates multiple foundation models.
\textcolor{Green}{Superscripts} denote the accuracy gains of POC over the corresponding FSL expert model (denoted ``exp'').
POC consistently improves all these methods.
\cref{fig:ablate_backbone} and \cref{fig:ablate_lmms} study POC with different backbones and LMMs, respectively.
Supplement \S\ref{sec:detailed_results} shows results in the 4- and 8-shot settings, where the same conclusions hold.
}
\vspace{-1mm}
\label{tab:improve_fsr}
\scalebox{0.9}{
\setlength{\tabcolsep}{0.68em}
\begin{tabular}{l c c c c c c c c cc c c }
\toprule

\multirow{2}{*}{FSL method} & 
\multicolumn{2}{c}{Aves} &
\multicolumn{2}{c}{Insecta} &
\multicolumn{2}{c}{Weeds} &
\multicolumn{2}{c}{Mollusca} &
\multicolumn{2}{c}{Fungi} &
\multicolumn{2}{c}{\cellcolor{col33}mean acc.} 
\\

\cmidrule(lr){2-3}\cmidrule(lr){4-5}\cmidrule(lr){6-7}\cmidrule(lr){8-9}\cmidrule(lr){10-11}\cmidrule(lr){12-13}

& exp & POC & exp & POC & exp & POC & exp & POC & exp & POC & \cellcolor{col33}exp & \cellcolor{col33}POC \\

\midrule

CoOp \cite{zhou2022learning} \textsubscript{IJCV'22}  & 48.2 & 62.0$^{\textcolor{Green}{+13.8}}$ & 38.2 & 51.4$^{\textcolor{Green}{+13.2}}$ & 69.9 & 83.7$^{\textcolor{Green}{+13.8}}$ & 44.5 & 59.7$^{\textcolor{Green}{+15.2}}$ & 8.9 & 13.0$^{\textcolor{Green}{+4.1}}$ & 
\cellcolor{col33}41.9 & \cellcolor{col33}53.9$^{\textcolor{Green}{+12.0}}$ \\

MaPLe \cite{maple}  \textsubscript{CVPR'23}  & 47.4 & 61.0$^{\textcolor{Green}{+13.6}}$ & 15.7 & 29.2$^{\textcolor{Green}{+13.5}}$ & 41.5 & 60.6$^{\textcolor{Green}{+19.1}}$ & 23.6 & 42.9$^{\textcolor{Green}{+19.3}}$ & 3.1 & \ 6.6$^{\textcolor{Green}{+3.5}}$ & 
\cellcolor{col33}26.3 & \cellcolor{col33}40.0$^{\textcolor{Green}{+13.7}}$ \\
\midrule

Tip-Adapter \cite{tipadapter} \textsubscript{ECCV'22}    & 47.7 & 59.6$^{\textcolor{Green}{+11.9}}$ & 14.6 & 28.3$^{\textcolor{Green}{+13.7}}$ & 53.1 & 71.7$^{\textcolor{Green}{+18.6}}$ & 27.3 & \ 44.7$^{\textcolor{Green}{+17.4}}$ & 3.2 & \ 7.4$^{\textcolor{Green}{+4.2}}$ & 
\cellcolor{col33}29.2 & \cellcolor{col33}42.3$^{\textcolor{Green}{+13.1}}$ \\

CLIP-Adapter \cite{clipadapter}  \textsubscript{IJCV'23}   & 50.9 & 63.6$^{\textcolor{Green}{+12.7}}$ & 38.8 & 56.3$^{\textcolor{Green}{+17.5}}$ & 68.4 & 82.0$^{\textcolor{Green}{+13.6}}$ & 48.4 & 57.4$^{\textcolor{Green}{+9.0}}$ & 6.3 & 10.8$^{\textcolor{Green}{+4.5}}$ & 
\cellcolor{col33}42.6 & \cellcolor{col33}54.0$^{\textcolor{Green}{+11.4}}$ \\

\midrule

Few-shot LP \cite{radford2021learning} \textsubscript{ICML'21}   & 53.8 & 67.7$^{\textcolor{Green}{+13.9}}$ & 54.7 & 63.0$^{\textcolor{Green}{+8.3}}$ & 69.7 & \ 85.2$^{\textcolor{Green}{+15.5}}$ & 56.8 & 63.9$^{\textcolor{Green}{+7.1}}$ & 20.0 & 22.7$^{\textcolor{Green}{+2.7}}$ & 
\cellcolor{col33}51.0 & \cellcolor{col33}60.5$^{\textcolor{Green}{+9.5}}$ \\

CMLP \cite{lin2023multimodality} \textsubscript{CVPR'23} & 53.3 & 66.0$^{\textcolor{Green}{+12.7}}$ & 57.3 & 66.6$^{\textcolor{Green}{+9.3}}$ & 75.1 & 84.3$^{\textcolor{Green}{+9.2}}$ & 58.2 & 65.2$^{\textcolor{Green}{+7.0}}$ & 21.8 & 25.3$^{\textcolor{Green}{+3.5}}$ & 
\cellcolor{col33}53.1 & \cellcolor{col33}61.5$^{\textcolor{Green}{+8.4}}$ \\

CLAP \cite{clap24} \textsubscript{CVPR'24}   & \underline{57.0} & \underline{67.8}$^{\textcolor{Green}{+10.8}}$ & \underline{63.1} & \textbf{70.8}$^{\textcolor{Green}{+7.7}}$ & \underline{76.9} & \underline{85.0}$^{\textcolor{Green}{+8.1}}$ & \underline{63.5} & \textbf{71.6}$^{\textcolor{Green}{+8.1}}$ & \underline{24.9} & \underline{28.2}$^{\textcolor{Green}{+3.3}}$ & 
\cellcolor{col33}\underline{57.1} & \cellcolor{col33}\underline{64.7}$^{\textcolor{Green}{+7.6}}$ \\

\midrule
FineR \cite{liu2024democratizing}  \textsubscript{ICLR'24}    & 47.5 & 62.5$^{\textcolor{Green}{+15.0}}$ & 32.8 & \ 47.1$^{\textcolor{Green}{+14.3}}$ & 65.0 & \ 82.4$^{\textcolor{Green}{+17.4}}$ & 47.7 & \ 58.7$^{\textcolor{Green}{+11.0}}$ & 8.9 & 13.9$^{\textcolor{Green}{+5.0}}$ & 
\cellcolor{col33}40.4 & \cellcolor{col33}\ 52.9$^{\textcolor{Green}{+12.5}}$ \\

\midrule

FS-FT \cite{liu2025few}   \textsubscript{CVPR'25}   & \textbf{58.2} & \textbf{69.4}$^{\textcolor{Green}{+11.2}}$ & \textbf{63.8} & \underline{70.5}$^{\textcolor{Green}{+6.7}}$ & \textbf{80.7} & \textbf{87.7}$^{\textcolor{Green}{+7.0}}$ & \textbf{63.6} & \underline{69.2}$^{\textcolor{Green}{+5.6}}$ & \textbf{29.9} & \textbf{31.1}$^{\textcolor{Green}{+1.2}}$ & 
\cellcolor{col33}\textbf{59.2} & \cellcolor{col33}\textbf{65.6}$^{\textcolor{Green}{+6.4}}$ \\
\bottomrule
\end{tabular}
}
\end{table*}

{\bf Remarks.} 
Prior work has explored augmenting prompts with visual attribute descriptions \cite{menon2022visual, parashar2023prompting} or taxonomic information \cite{wu2024protect}.
However, we find that such additions offer no benefits beyond our prompting strategy (\cref{fig:mmICL-prompt}) and can even degrade performance (detailed in Supplement \S\ref{sec:detailed_results}).
This observation is consistent with recent studies \cite{parashar2023prompting, saha2024improved}.
We conjecture that this is because 
(1) the images in the prompt already provide sufficient evidence for the LMM to reason about species, making additional descriptions unnecessary,
and (2) LMMs struggle to ground textual descriptions to fine-grained visual regions \cite{sun2024alpha,wei2023ov, kim2026vlm}, causing extra textual information to hinder rather than facilitate reasoning.
In sum, 
POC effectively combines the complementary strengths of the FSL expert model and the LMM.
It is training-free, validation-free, and model-agnostic, requiring no manual intervention.
These properties make POC a general and practical framework for VSR,
readily compatible with existing FSL approaches and LMMs, as shown in \cref{tab:improve_fsr}, Figs.~\ref{fig:ablate_backbone}-\ref{fig:ablate_lmms}.

\section{Experiments}
\label{sec:experiments}

We conduct extensive experiments to validate our POC, demonstrating its superior performance over compared methods and compatibility with various FSL expert models and LMMs.
We first describe the experimental setup, including compared methods, models, and implementation details.
We then present comprehensive results and analyses.

\subsection{Experimental Setup}
\label{ssec:exp_setup}

{\bf Compared Methods.}
We compare against several categories of FSL approaches,
including 
prompt learning (CoOp \cite{zhou2022learning} and MaPLe \cite{maple}),
adapter learning (Tip-Adapter \cite{tipadapter} and CLIP-Adapter \cite{clipadapter}),
classifier learning (Few-shot LP \cite{radford2021learning}, CMLP \cite{lin2023multimodality}, and CLAP \cite{clap24}),
and few-shot finetuning \cite{liu2025few}.
Regarding LMM-based approaches,
we compare with FineR \cite{liu2024democratizing}, a dedicated fine-grained recognition method that integrates multiple FMs.
Notably, rather than using the original FineR that was designed for the zero-shot setting,
we enhance it by leveraging the few-shot labeled training images (\cref{fig:annotaion-example}).
Specifically, FineR predicts the species of a test image by (1) first using an LMM to describe the visual attributes of unlabeled images and then using an LLM to infer candidate species names, (2) computing the similarity between the test image and the visual and text attributes of the candidate species using a VLM, (3) assigning the closest class as the final prediction.  
In contrast, we directly exploit the labeled training images by (1) matching the test image to the training images of each class in the feature space, and (2) assigning the ground-truth label of the nearest class center as the prediction.

\textbf{Models.}
As most recent FSL approaches are built upon pretrained VLMs,
we follow prior work \cite{liu2025few} and adopt the open-source OpenCLIP ViT-B/32 \cite{cherti2023reproducible} as the backbone for all FSL approaches in our benchmarking experiments.
To further evaluate the generality of POC,
we consider several alternative visual backbones,
including ResNet50 models pretrained on ImageNet \cite{deng2009imagenet} and iNaturalist \cite{su2021realistic}, and DINOv2 ViT-B/14 \cite{oquab2023dinov2}. 
For LMM,
we use the open-source Qwen-2.5-VL-7B-Instruct \cite{qwen2.5-vl} as the default model for benchmarking, and additionally evaluate the GLM-4.1V-9B-Thinking \cite{vteam2025glm} and GPT-5-Mini \cite{openai2025gpt5} to demonstrate the compatibility of POC across diverse LMMs.

{\bf Implementations.}
To train expert models using FSL methods,
we adopt the default hyperparameters reported by each FSL method (detailed in Supplement \S\ref{sec:hyperparams}).
When comparing different LMMs within our POC, 
we use the same prompt template (\cref{fig:mmICL-prompt}) throughout our experiments.
For open-source LMMs,
we host them on a single NVIDIA A100 GPU (40GB).
As a reference for computational cost, 
running POC with Qwen-2.5-VL-7B-Instruct on the 4,000 test images of Aves requires $\sim$6 GPU hours, corresponding to $\sim$5 GPU seconds per image.
For GPT-5-Mini \cite{openai2025gpt5},
which is accessed through the API, the total cost for all five datasets is below \$50.
Furthermore, 
since most recent FSL methods report results under the 16-shot setting \cite{liu2025few, clap24, maple}, 
we adopt the same setting in the main paper,
and provide experiments under the 4- and 8-shot settings in Supplement \S\ref{sec:detailed_results},
where POC consistently delivers strong performance gains.
We release our code at \url{https://tian1327.github.io/POC}
for the community to reproduce our results.

\begin{figure}[!t]
\centering
\vspace{-5mm}
\includegraphics[width=0.99\linewidth, clip=true, trim=0mm 0mm 50mm 0mm]{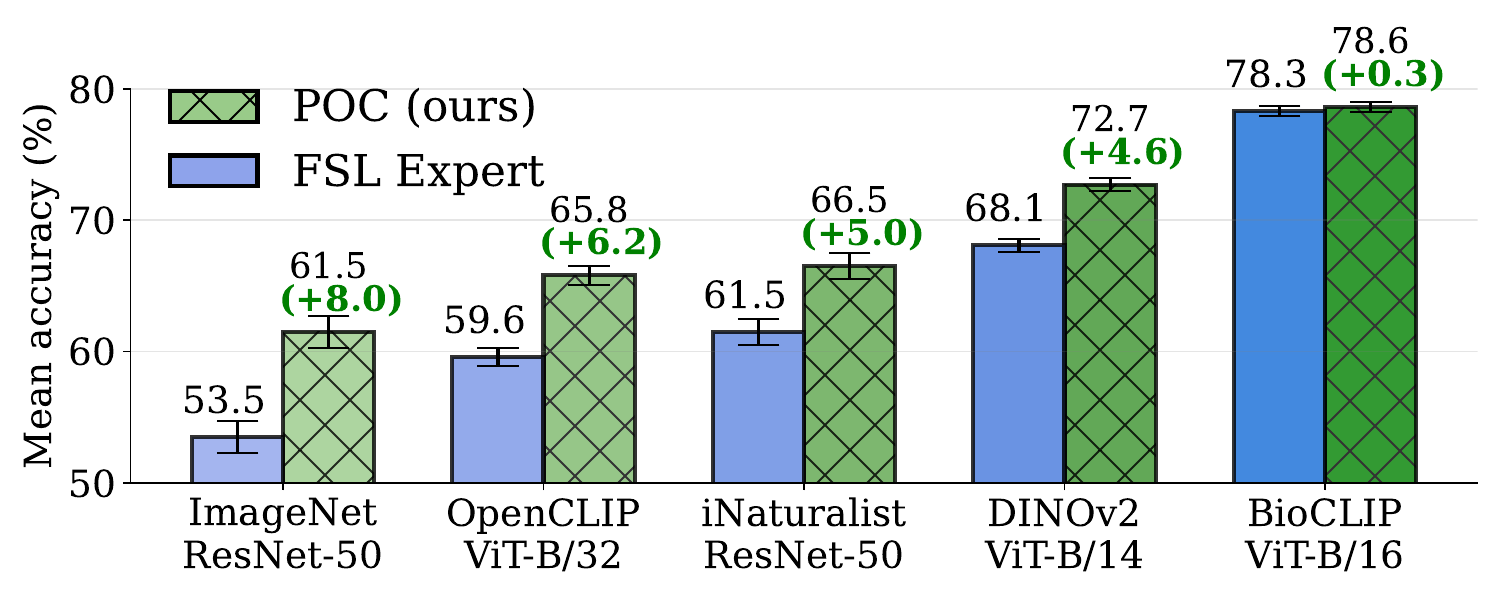}
\vspace{-3mm}
\caption{\small
\textbf{POC generalizes across different FSL expert models.}
Following \cite{liu2025few}, we finetune each backbone in the 16-shot setting to obtain FSL experts.
We then apply POC with the LMM Qwen-2.5-VL-7B-Instruct. 
For each FSL expert model, 
we report its mean accuracy across five benchmark datasets over three runs with different few-shot data sampled with three random seeds - 1, 2, and 3.
POC consistently improves all expert models, 
yielding higher accuracy for stronger expert models and larger gains for weaker ones.
}
\label{fig:ablate_backbone}
\vspace{-1mm}
\end{figure}

\subsection{Results}
{\em POC significantly improves existing FSL methods for VSR.}
\cref{tab:improve_fsr} highlights the advantages of POC in serving as a  plug-in module that 
greatly enhances diverse FSL methods.
In contrast to the poor performance of LMMs when prompted with open-vocabulary questions (\cref{tab:lmm_failue}), POC's significant improvements  
confirm our successful exploitation of LMM for VSR. 
The enhanced FineR obtains 40.4\% mean accuracy, 
while POC improves it to 52.9\%. 
In addition, 
POC improves the strong expert model learned by FS-FT \cite{liu2025few} from 59.2\% to 65.6\%.
All the results validate the complementary strengths of FSL expert models and LMMs,
and that POC can leverage both to enhance VSR.

{\em POC generalizes across different  FSL expert backbones.}
\cref{fig:ablate_backbone} shows that POC yields consistent accuracy gains over expert models that finetune various pretrained backbones via FS-FT~\cite{liu2025few}. 
Moreover, POC achieves higher accuracy with stronger expert models
and yields larger gains for weaker ones.
This is easy to understand -- 
with a fixed LMM,
the final performance of POC is upper-bounded by the top-5 accuracy of the expert models, while the top-5 accuracy is closely correlated with the top-1 accuracy, as is widely known \cite{russakovsky2015imagenet}.
It is also worth noting that the DINOv2 ViT-B/14 backbone is pretrained on general images in a self-supervised fashion.
In contrast, the iNaturalist ResNet-50 backbone is pretrained in a supervised learning manner over 438K meticulously curated bio-images labeled by either domain experts or community efforts \cite{van2018inaturalist}.
Yet, the superior performance of DINOv2 over iNaturalist-ResNet-50 suggests that DINOv2 is a more practical backbone in VSR over customized bio-models.

\begin{figure}[!t]
\vspace{-2mm}
\centering
\includegraphics[width=\linewidth]{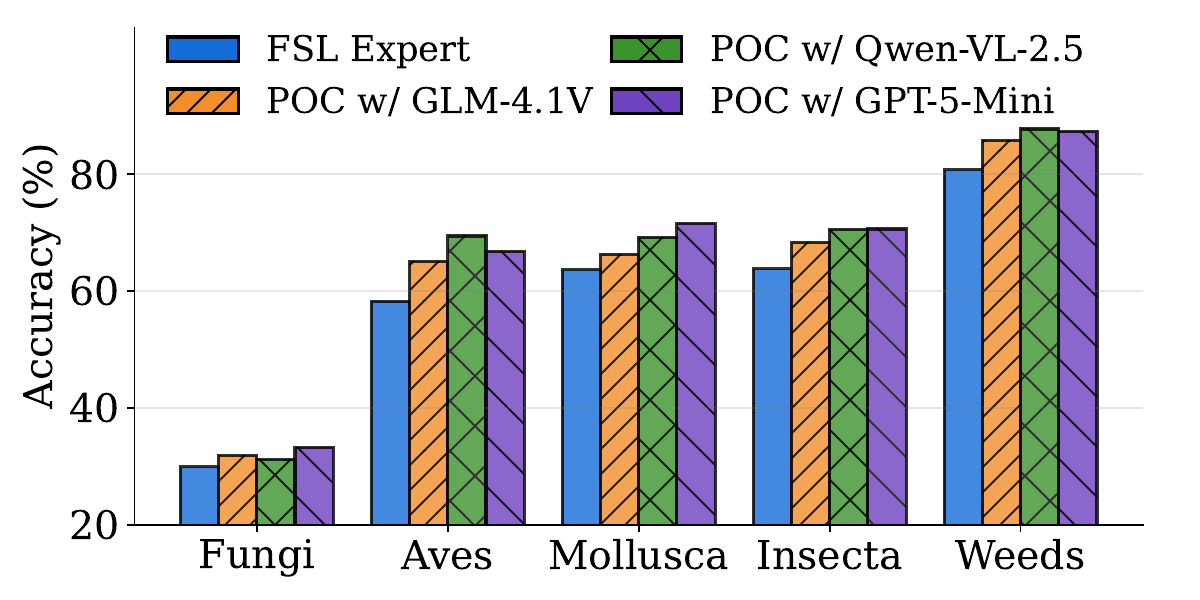}
\vspace{-6mm}
\caption{\small
\textbf{POC generalizes across different LMMs.}
Using the FSL expert model by few-shot finetuning  \cite{liu2025few} the OpenCLIP ViT-B/32 backbone in the 16-shot setting, 
we evaluate POC using different LMMs, 
including Qwen-2.5-VL-7B-Instruct \cite{qwen2.5-vl}, 
GLM-4.1V-9B-Thinking \cite{vteam2025glm}, 
and GPT-5-Mini \cite{openai2025gpt5}. 
The consistent gains over the expert model across LMMs demonstrate that POC can effectively leverage the reasoning capabilities of LMMs to enhance VSR.
}
\label{fig:ablate_lmms}
\vspace{-2mm}
\end{figure}

\begin{figure}[t]
    \centering
    \includegraphics[width=\linewidth, clip=true, trim=0mm 0mm 0mm 0mm]{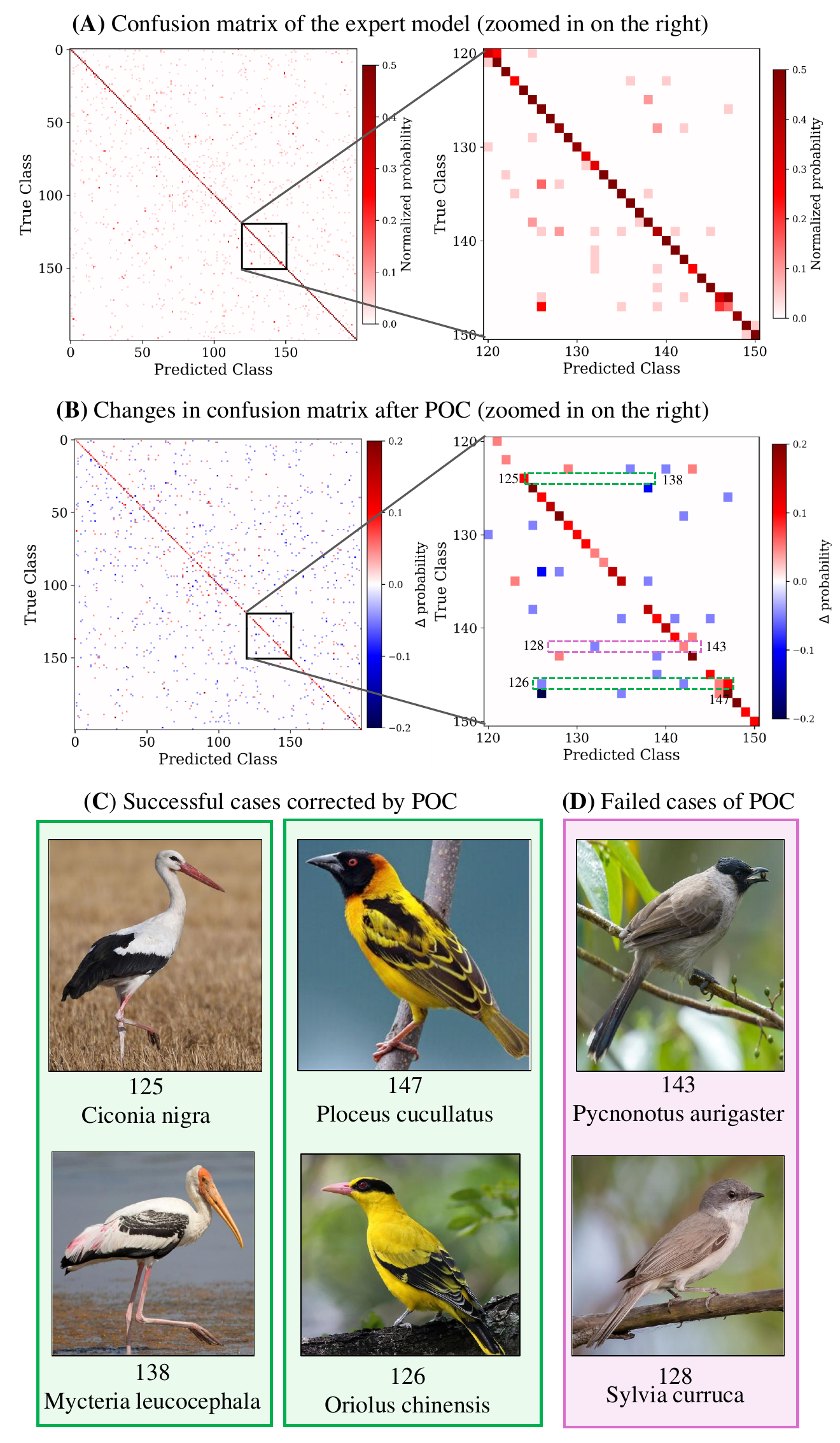}
    \vspace{-6.5mm}
    \caption{\small
    \small
\textbf{(A)} depicts the confusion matrix of the FSL expert \cite{liu2025few} on the test data of the Aves benchmark (200 classes).
The diagonal and off-diagonal dots denote the accuracies on individual species and misclassification rates for certain species pairs. 
\textbf{(B)} depicts the changes in accuracy and misclassification rates after applying POC,
where \textcolor{Red}{red} diagonal dots indicate improved accuracy and \textcolor{blue}{blue} off-diagonal dots indicate reduced misclassification rates. Performance gains are reflected by the red diagonals and blue off-diagonals.
\textbf{(C)} shows some examples of confusing species pairs, previously misclassified by the expert model but corrected by POC. Note the subtle differences in the beak shape and plumage.
\textbf{(D)} shows failure cases where POC occasionally makes mistakes on test images that were originally correctly identified by the expert model
Additional examples from other benchmarks are shown in Supplement \cref{sec:detailed_results}.
}
\vspace{-4mm}
\label{fig:confusion_matrix}
\end{figure}

\begin{figure}[t]
\centering
\vspace{-1.5mm}
\includegraphics[width=\linewidth]{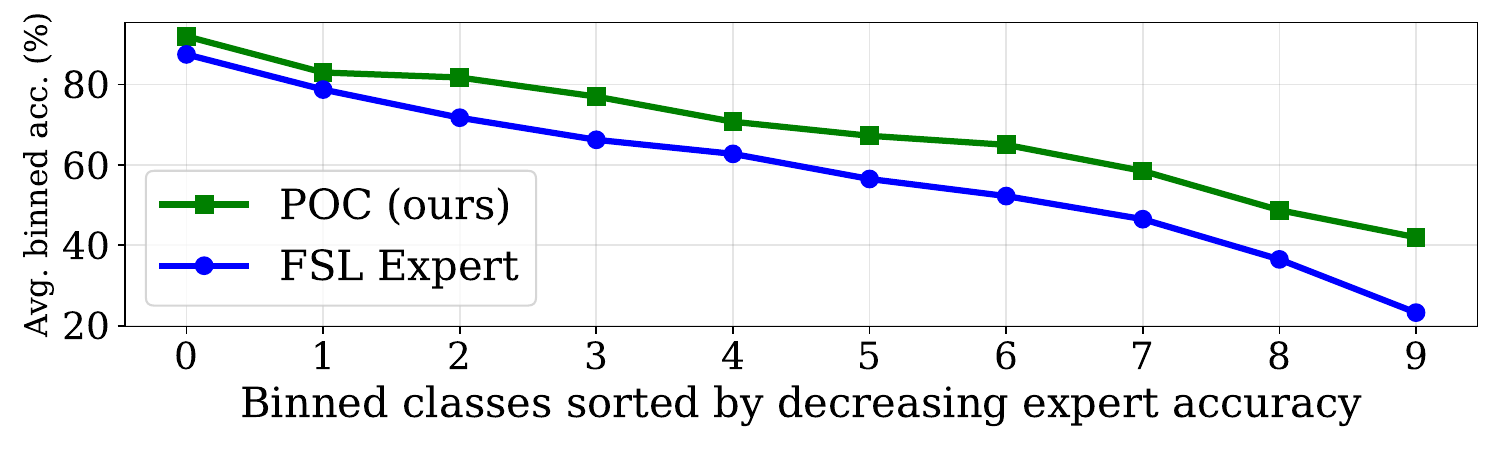}
\vspace{-7mm}
\caption{
\small
\textbf{Breakdown analysis of POC's improvements on easy and hard species.}
Species in the Aves benchmark are ranked by the per-species accuracy of the FSL expert model.
We plot the corresponding accuracies after applying POC, grouping species into bins to reduce noise and reveal the overall trend -- POC generally yields larger accuracy gains on harder species.
Supplement \S\ref{sec:detailed_results} displays similar plots on other benchmarks.
}
\vspace{-2mm}
\label{fig:improve_acc}      
\end{figure}

{\em POC generalizes across different LMMs.}
As shown in \cref{fig:ablate_lmms},
POC consistently improves the performance of an FSL expert model when paired with different LMMs,
demonstrating its broad compatibility. 
Furthermore, 
stronger LMMs, such as GPT-5-Mini \cite{openai2025gpt5}
and Qwen-VL-2.5-Instruct \cite{qwen2.5-vl},
yield larger gains,
suggesting that POC effectively exploits the superior reasoning capabilities of more powerful LMMs.

{\em POC improves FSL expert models by distinguishing confusing species.}
\cref{fig:confusion_matrix} studies when POC succeeds and fails.
In particular, \cref{fig:confusion_matrix}A plots the confusion matrix of the FSL expert model,
and \cref{fig:confusion_matrix}B plots the changes in the confusion matrix after applying POC.
By zooming in on the classes and visualizing corresponding images 
(\cref{fig:confusion_matrix}C),
results show that the FSL expert model does not perform well in distinguishing between pairs of bird species that exhibit highly similar visual appearances. 
However, 
POC can often correct these misclassifications while occasionally introducing new mistakes (\cref{fig:confusion_matrix}D).
Moreover, 
\cref{fig:improve_acc} shows that POC improves accuracy for both easy and hard species and generally yields larger gains for harder ones.

\begin{figure}[t]
\centering
\vspace{-1.5mm}
\includegraphics[width=\linewidth]{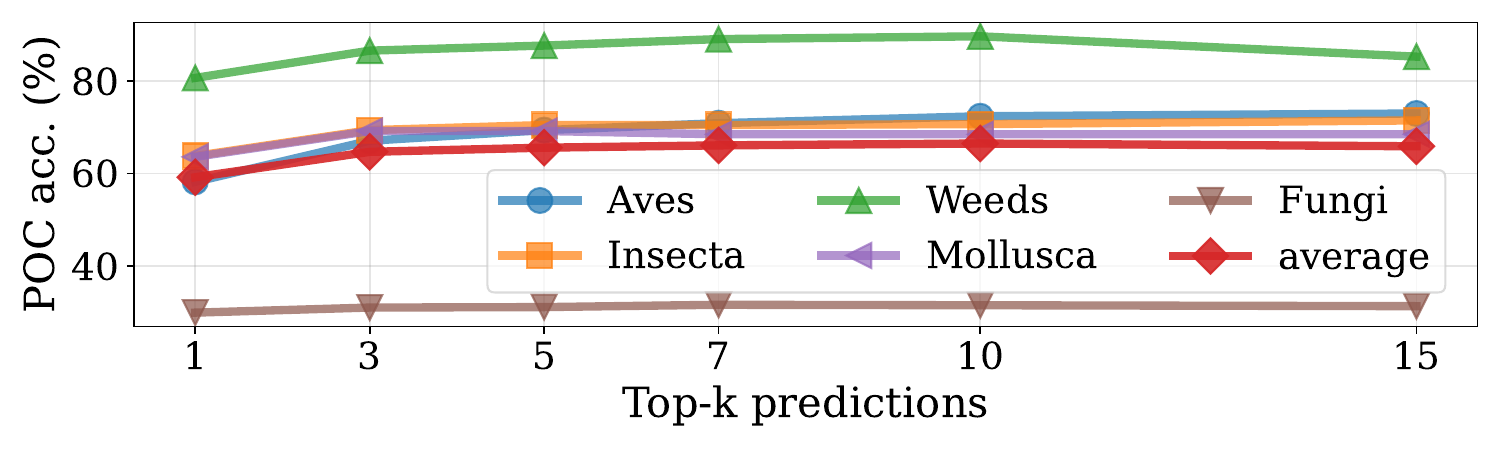}
\vspace{-7mm}
\caption{
\small
\textbf{Effects by varying $k$ when prompting with top-$k$ candidate species.}
With the same FSL expert model and LMM as those in \cref{tab:improve_fsr}, 
we evaluate POC by prompting with different numbers of top-$k$ predicted species from the expert model.
POC yields higher accuracy with larger $k$ but saturates around $k=10$ for all the datasets.
}
\label{fig:ablate_topk}
\end{figure}

{\em Impact of varying $k$ for prompting top-$k$ candidate species.}
\cref{fig:ablate_topk} shows that including more top predictions from the FSL expert model is generally beneficial, as it increases the likelihood that candidates contain the true species.
However, 
the performance gains plateau beyond around $k=10$ across all datasets.

{\em Generalization of POC beyond VSR.}
Although the focus of our work is VSR,
we further evaluate POC on several popular fine-grained benchmarks outside the biological domain, 
including
FGVC Aircraft, DTD, Food-101, and Stanford Cars.
The Supplemental Table~\ref{tab:nonvsr_generalization} shows that POC consistently improves the few-shot finetuned expert model on these datasets,
improving their mean accuracy from 70.3\% to 77.3\%.
This demonstrates the generality of our POC for broad fine-grained recognition tasks.

\textit{Analysis of foundation models specialized for bio-image analysis.}
Motivated by the importance of VSR, 
recent studies have developed FMs specialized for bio-image recognition,
such as BioCLIP \cite{stevens2024bioclip}.
These models are trained on tens of millions of labeled bio-images,
collected from existing datasets that are annotated by experts and citizen scientists,
reflecting a significant prior annotation effort.
Therefore, direct comparison with such models falls outside the scope of our focused few-shot setting (\S\ref{sec:problem}).
Nevertheless, for completeness, we evaluate the performance of the BioCLIP ViT-B/16 model.
To maximize its performance in an ``unfairly favorable'' way, 
we train an expert model by finetuning \cite{liu2025few} BioCLIP on the few-shot training images provided in each benchmark.
We then compare this expert model against SimpleShot \cite{wang2019simpleshot}, which is built on the original BioCLIP (the FSL method used in the BioCLIP paper \cite{stevens2024bioclip}), and against our POC framework.
Table~\ref{tab:bioclip_combined} summarizes the results on the five VSR benchmarks under the 16-shot setting, and on BioCLIP's three test sets under its recommended 5-shot setting.
On the VSR benchmarks, 
SimpleShot is already strong, surpassing all the methods studied in Table~\ref{tab:improve_fsr};
the expert model yields decent gains over it, but POC barely improves the expert further.
The strong performance of the FSL methods and the limited gains from POC can be attributed to the data leakage. 
For example, BioCLIP was pretrained on iNaturalist images, from which our VSR benchmarks are constructed.
In contrast, on BioCLIP's test sets,
POC yields notable gains, 
highlighting its practical value in specialized domains, where large-scale labeled data are unavailable or inaccessible for foundation model pretraining.

\section{Impacts and Limitations}
\label{sec:discussion}

{\bf Impacts.} 
Our work advances VSR,
with potential benefits for a broad range of disciplines, including evolutionary biology, ecology, environmental science, biodiversity conservation, and climate-change research.
The simplicity and model-agnostic nature of our approach lower the barrier to adopting advanced AI techniques,
making the state-of-the-art visual recognition tools more accessible to scientific communities that may lack the resources or expertise to develop and deploy large-scale AI systems.
Regarding potential negative impact,
the performance of our POC framework ultimately depends on both the underlying expert model and the LMM,
either of which may produce erroneous predictions.
Consequently, we emphasize that POC should be thoroughly evaluated before deployment in domain-specific applications.
For instance, in ecological monitoring and biodiversity conservation, misidentifying invasive species or endangered species could lead to incorrect management decisions and potentially adverse environmental consequences. 
Therefore, human oversight and rigorous validation remain essential when applying POC in high-stakes real-world settings.

\begin{table}[t]
\centering
\small
\caption{\small 
\textbf{POC with BioCLIP-based expert model.}
Using the BioCLIP ViT-B/16 backbone, we adopt SimpleShot \cite{wang2019simpleshot}, the FSL method used in the original BioCLIP paper, 
and construct a stronger expert model by finetuning \cite{liu2025few} the backbone under the 16-shot setting on our five VSR benchmarks and the 5-shot setting used by BioCLIP on its three evaluation datasets.
On our benchmarks, SimpleShot already achieves strong performance, surpassing all methods compared in Table~\ref{tab:improve_fsr}; 
the expert model decently outperforms it, but POC barely improves further.
As discussed in the main text, the strong performance of FSL methods and the limited gains of POC are likely due to data leakage.
However, on BioCLIP's test sets, POC yields notable gains, validating the effectiveness of our POC framework. Superscripts indicate the accuracy gains or degradations of POC relative to the corresponding FSL expert model.
}
\label{tab:bioclip_combined}
\vspace{-2mm}
\scalebox{0.8}{
\setlength{\tabcolsep}{0.38em}
\begin{tabular}{llllll>{\columncolor{col33}}l}
\toprule
\multicolumn{7}{c}{\textbf{Test accuracy on the five VSR benchmarks}} \\
\midrule
Method 
& Aves 
& Insecta 
& Weeds 
& Mollusca 
& Fungi 
& \cellcolor{col33}mean acc. \\

\midrule
SimpleShot \cite{wang2019simpleshot} 
& 76.1 & 81.1 & 94.9 & \textbf{82.2} & 39.6 & \cellcolor{col33}74.8\\

FSL expert \cite{liu2025few}
& 83.8 & 86.5 & 97.0 & 79.3 & \textbf{44.9}
& \cellcolor{col33}78.3 \\

\textbf{POC} (ours)
& \textbf{85.5}$^{\textcolor{Green}{+1.7}}$
& \textbf{86.5}$^{\textcolor{Green}{+0.0}}$
& \textbf{97.2}$^{\textcolor{Green}{+0.2}}$
& 79.0$^{\textcolor{red}{-0.3}}$
& 44.7$^{\textcolor{red}{-0.2}}$
& \cellcolor{col33}\textbf{78.6}$^{\textcolor{Green}{+0.3}}$ \\
\midrule
\midrule
\end{tabular}
}

\scalebox{0.8}{
\setlength{\tabcolsep}{0.50em}
\begin{tabular}{llll>{\columncolor{col33}}l}
\multicolumn{5}{c}{\textbf{Test accuracy on BioCLIP's challenging test sets}} \\
\midrule
Method 
& RareSpecies \cite{stevens2024bioclip}
& Insect2 \cite{meta-album-2022} 
& PlantDoc \cite{meta-album-2022}
& \cellcolor{col33}mean acc. \\

\midrule

SimpleShot \cite{wang2019simpleshot}
& 65.7  & 33.6 &  47.5
& \cellcolor{col33}48.9 \\

FSL expert \cite{liu2025few}
& 70.8 &  38.4 & 48.1
& \cellcolor{col33}52.4 \\

\textbf{POC} (ours)
& \textbf{74.5}$^{\textcolor{Green}{+3.7}}$
& \textbf{39.3}$^{\textcolor{Green}{+0.9}}$
& \textbf{50.2}$^{\textcolor{Green}{+2.1}}$
& \cellcolor{col33}\textbf{54.7}$^{\textcolor{Green}{+2.3}}$ \\
\bottomrule
\end{tabular}
}
\vspace{-2mm}

\end{table}

{\bf Limitations and Future Work.}
We discuss two notable limitations.
First, although our problem formulation targets image-based species recognition and is applicable to a wide range of scenarios, identifying certain species requires ``invisible cues'' beyond conventional RGB imagery.
For example, to materialize such invisible cues,
UV imaging is used to identify bird and fish species \cite{bleiweiss2004ultraviolet, losey1999uv},
and high-resolution microscopy is adopted for pollen species identification  \cite{romero2020improving}.
Such specialized imaging modalities are rarely represented in publicly available datasets and hence are unlikely to be learned well by current FMs and LMMs.
Investigating these challenging settings is an important direction for future research.
Second, our POC framework invokes an LMM after an FSL expert model, introducing additional inference latency (e.g., $\sim$5s per test image).
While real-time inference is not a strict requirement in many scientific applications (e.g., biodiversity monitoring systems often process surveillance imagery or video data offline),
the overhead may limit the deployment in latency-sensitive applications such as smartphone-based species identification.
Nevertheless, with more efficient LMMs and inference techniques, we expect the computational cost of POC to decrease substantially.

\section{Conclusions}
\label{sec:conclusions}

We study Visual Species Recognition (VSR) from first principles,
motivated by the real-world annotation practice,
where domain experts craft an instruction by using a few images and text descriptions to specify each species of interest.
This leads to a multimodal few-shot learning (FSL) setup.
To address VSR,
we systematically compare two distinct paradigms:
(1) training expert models using FSL,
and (2) prompting advanced Large Multimodal Models (LMMs) for VSR.
We find that LMM-based recognition significantly underperforms FSL expert models,
despite extensive prompt engineering.
Interestingly,
we observe that LMMs can frequently correct wrong predictions of an FSL expert model when provided with the expert's top-$k$ prediction.
Hence, we are motivated to propose the Post-hoc Correction (POC) framework,
which leverages an LMM to refine the predictions of an FSL expert model.
POC adopts a multimodal prompting strategy that incorporates the expert model's top-$k$ predicted species, their confidence scores, and the corresponding few-shot training images.
In extensive experiments,
POC consistently improves VSR performance across different datasets, FSL methods, backbones, and LMMs.
POC is training-free, validation-free, and model-agnostic, making it a strong baseline in real-world VSR tasks.

\section*{Acknowledgement}
This work was supported by the Science and Technology Development Fund of Macau (0058/2025/RIA2, 0067/2024/ITP2), the University of Macau (SRG2023-00044-FST), the Institute of Collaborative Innovation, and the CK Foundation.
TL and AB were supported by the CSE Department at Texas A\&M University.
This work utilized the advanced computing resources and consultation provided by Texas A\&M High Performance Research Computing (HPRC), and the Delta system at the National Center for Supercomputing Applications [award OAC 2005572] through allocation [CIS250837, CIS250928] from the Advanced Cyberinfrastructure Coordination Ecosystem: Services \& Support (ACCESS) program, which is supported by National Science Foundation grants \#2138259, \#2138286, \#2138307, \#2137603, and \#2138296.

\ifCLASSOPTIONcaptionsoff
  \newpage
\fi



\bibliographystyle{IEEEtran}
\bibliography{PAMI/egbib-new-restored}

%


\vspace{-1cm}
\begin{IEEEbiography}
[{\includegraphics[width=1in,height=1.15in,clip,keepaspectratio]{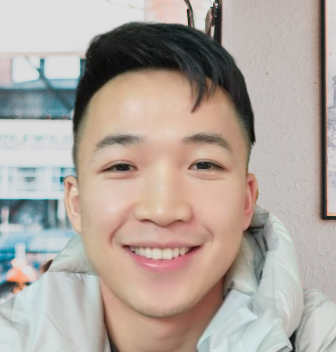}}]
{Tian Liu} received the Ph.D. degree in computer science at Texas A\&M University. 
His research interests include computer vision, cyber-physical systems, and machine learning for healthcare. His current research focuses on Vision-Language Models, especially on developing better learning methods for adapting pretrained Vision-Language Models for downstream tasks in the zero-shot and few-shot setups.
\vspace{-1.5cm}
\end{IEEEbiography}


\begin{IEEEbiography}[{\includegraphics[width=1in,height=1.25in,clip,keepaspectratio]{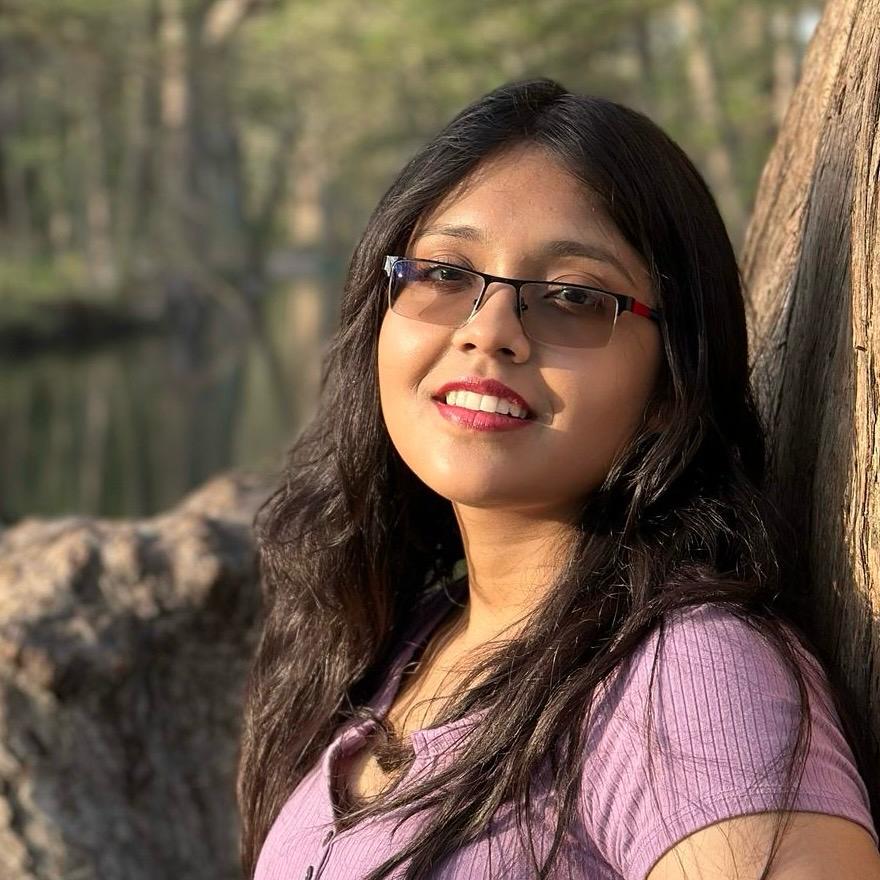}}]{Anwesha Basu} is a Ph.D. student in the Department of Computer Science and Engineering at Texas A\&M University, where she earned the M.S. degree in Computer Science.
Her research interests include computer vision, multimodal AI, vision-language models, and multimodal retrieval.
Her current research focuses on large-scale multimodal retrieval and representation learning systems.
\vspace{-1cm}
\end{IEEEbiography}


\begin{IEEEbiography}
[
{\includegraphics[width=1in,height=1.25in,clip,keepaspectratio]{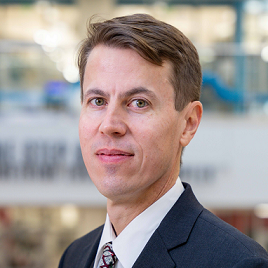}}]{James Caverlee}
(Member, IEEE) received the B.A.
degree (magna cum laude) in economics from Duke
University in 1996,
the M.S.
degrees in engineering-economic systems and operations research and computer science from Stanford University in 2000 and
2001, respectively, and the Ph.D. degree in computer
science from Georgia Tech in 2007.
He is a Professor and Lynn'84 and Bill Crane'83 Faculty Fellow with the Department of Computer Science and Engineering, Texas A\&M University.
Dr. Caverlee was a recipient of the NSF CAREER Award, the DARPA
Young Faculty Award, and the AFOSR Young Investigator Award, as well as
several teaching awards. He was the Conference General Co-Chair for WSDM
2020.
\vspace{-1cm}
\end{IEEEbiography}


\begin{IEEEbiography}[{\includegraphics[width=1in,height=1.25in,clip,keepaspectratio]{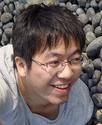}}]{Shu Kong} is an Assistant Professor of Computer Science at the University of Macau.
He was an Assistant Professor in the Department of Computer Science and Engineering at Texas A\&M University.
He held positions as a Project Scientist and Postdoctoral Fellow in the Robotics Institute at Carnegie Mellon University. He received his Ph.D. in Computer Science from the University of California, Irvine.
His research aims to establish the foundations of Visual Intelligence.
To advance this vision,
he established the research paradigm Open World Vision, on which his paper was recognized for Best Paper / Marr Prize at ICCV 2021.
Beyond developing fundamental algorithms, he actively applies them to high-impact scientific domains, such as palynology, paleoecology, paleontology, and long-term evolutionary studies.
\end{IEEEbiography}




\clearpage

\renewcommand{\thesection}{\Alph{section}}
\renewcommand{\theHsection}{\Alph{section}}
\setcounter{section}{0}

\twocolumn[
  \begin{center}
    {\Huge \sffamily 
    Visual Species Recognition with Large Multimodal Models as Post-Hoc Correctors
    \\ 
    \vspace{4mm}
    \emph{(Supplemental Document)}
    } \\ 
    \vspace{1em}
    {\large Tian Liu$^{*}$,
Anwesha Basu$^{*}$,
James Caverlee,
Shu Kong$^{\dag}$}
  \end{center}
  \vspace{2em} 
]


\renewcommand{\thesection}{\Alph{section}}
\setcounter{section}{0} 
\section*{}
\begin{center}
    \emph{\bf \em \large Outline}
\end{center}
{This document supplements the main paper with additional information. The outline below summarizes the document.

\begin{itemize}[topsep=-3pt, partopsep=3pt, leftmargin=0.3in]

\item {\bf Section \ref{sec:datasets}} describes the statistics of  benchmarking datasets. 

\item {\bf Section \ref{sec:hyperparams}} provides hyperparameters in model training.

\item {\bf Section \ref{sec:Demo-code}} provides code and instructions for replicating our results.

\item {\bf Section~\ref{sec:prompt_templates}} presnets various prompt templates of POC.

\item {\bf Section \ref{sec:detailed_results}} 
reports per-dataset results in experiments,  additional visual examples, and confusion matrices.

\end{itemize}
}

\section{Summary of Datasets}
\label{sec:datasets}

\cref{tab:datasets} summarizes the details of the benchmarks used in our study. We obtain the five main VSR benchmarks from three large-scale publicly available biological datasets, including iNaturalist~\cite{inat2021}, Species196~\cite{he2023species196}, and FungiTastic~\cite{picek2025fungitastic}. Specifically, we keep the 200 classes from the Aves subset~\cite{semi-aves} of iNaturalist for the Aves benchmark. 
From Species196~\cite{he2023species196}, we sample classes that contain at least 20 labeled images, allowing 16-shot training and at least 4  remaining images for reliable evaluation.
This results in 78 classes for Insecta, 20 classes for Weeds, and 7 classes for Mollusca. From the training images of the selected classes, we randomly sample $K$-shot ($K = 4, 8, 16$) labeled data with three random seeds 1, 2, and 3 for our few-shot learning setup. We evaluate all methods on a held-out test set for each benchmark. \cref{fig:handpicked_examples} shows visual examples from each VSR benchmark. 

In the supplement, we also evaluate POC on BioCLIP test sets \cite{stevens2024bioclip}, including RareSpecies \cite{stevens2024bioclip}, Insect2 \cite{meta-album-2022}, and PlantDoc \cite{meta-album-2022}. We follow the same 5-shot setting for a fair comparison with results reported in \cite{stevens2024bioclip}.
Furthermore, to assess the generalization of POC to other domains, we evaluate it on popular non-VSR datasets, including FGVC Aircraft \cite{aircraft}, Stanford Cars \cite{cars}, Describable Textures Dataset (DTD) \cite{dtd}, and Food-101 \cite{food}.  
Table~\ref{tab:datasets} summarizes the statistics of all the datasets.

{
\setlength{\tabcolsep}{0.6em}
\begin{table}[ht]
\centering
\small
\caption{\small \textbf{Details of benchmarks used in our evaluations.}
We list the number of classes, original training data, and test data. 
Our few-shot setup randomly samples $K$-shot ($K$ = 4, 8, 16) images per class from the original training set to learn expert models.
}
\vspace{-3mm}
\label{tab:datasets}

\scalebox{0.85}{
\begin{tabular}{llrrrr}
\toprule 
type & dataset & \# class   &  \# original train set & \# test set \\
\midrule
\multirow{5}{*}{\makecell{VSR\\(ours)}}
& Aves & 200 & 5,959 & 4,000 \\
& Insecta  & 78 & 2,960 & 2,910 \\
& Weeds  & 20 & 890 & 913 \\
& Mollusca  & 7 & 688 & 704 \\
& Fungi  & 196 & 61,928 & 3,805 \\
\midrule
\multirow{3}{*}{\makecell{BioCLIP\\test sets}}
& RareSpecies \cite{stevens2024bioclip}  & 400 & 4,923 & 3,950 \\
& Insects2 \cite{meta-album-2022}  & 102 & 1,632 & 2,448 \\
& PlantDoc \cite{meta-album-2022}  & 27 & 432 & 648 \\
\midrule
\multirow{4}{*}{non-VSR}
& Aircraft \cite{aircraft}  & 100 & 3,334 & 3,333 \\
& DTD \cite{dtd}& 47 & 2,820 & 1,692 \\
& Food101 \cite{food}  & 101 & 50,500 & 4,040 \\
& StanfordCars \cite{cars}  & 196 & 6,509 & 3,920 \\
\bottomrule
\end{tabular}}
\end{table}
}

\section{Hyperparameters}
\label{sec:hyperparams}

Our POC is training-free, validation-free, and requires no human intervention. The only hyperparameter is $k$, representing the number of top predictions from the expert model that are used to prompt the LMM. We set $k$ to 5 by default in our experiments.
We postpone the analysis of $k$ Section \ref{sec:detailed_results} (ref. \cref{tab:ablate_topk_detail} therein).
Below, we detail the hyperparameters for learning the few-shot expert models.

\begin{figure*}[!ht]
  \centering
  \includegraphics[width=0.99\linewidth, trim=0mm 0mm 0mm 0mm, clip]{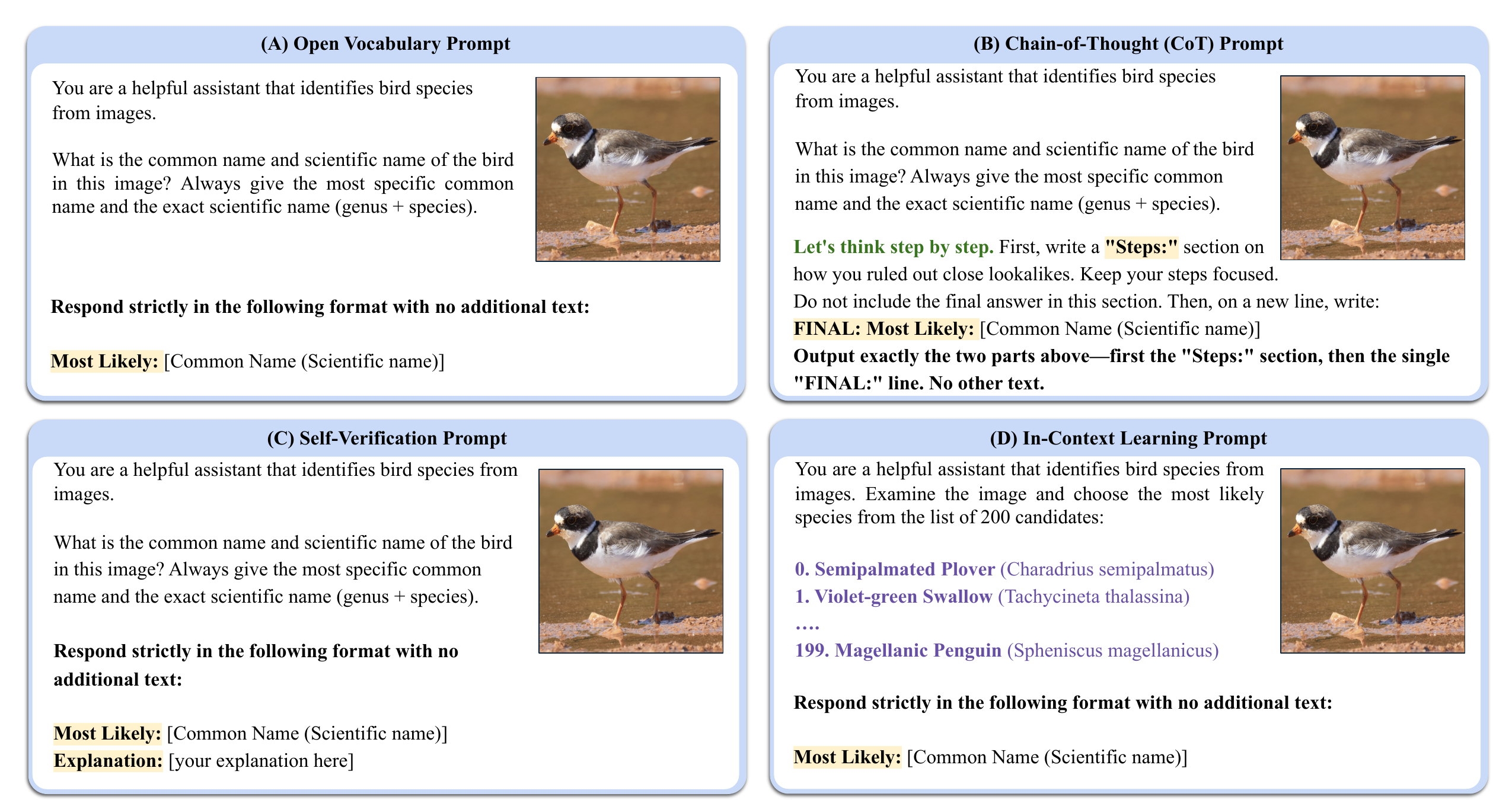}
  \caption{ \small \textbf{Examples of various prompt templates used on Aves benchmark}. Given a test image, we prompt an LMM asking it to predict the species names in open vocabulary (A), further enhanced by asking it to output the reasoning steps through Chain-of-Thought (B) or conduct self-verification by explaining its decisions (C). We also explore in-context learning by providing the complete list of class names (D).
  \cref{tab:lmm_poc_summary} highlights that LMMs struggle with VSR, even with these established prompting techniques.
  In contrast, our POC adopts a multimodal prompt (shown in \cref{fig:mmICL-prompt}), achieving significant gains.
  }
  \label{fig:pt_templates}
\end{figure*}

{\bf Few-shot-FT expert.}
Following the established validation-free few-shot learning protocol \cite{liu2025few}, we directly adopt the hyperparameters and the few-shot learning pipeline reported in \cite{liu2025few}.
Specifically, for foundational VLMs such as OpenCLIP \cite{cherti2023reproducible} and BioCLIP \cite{stevens2024bioclip}, it first initializes the classifier weights using the text embeddings of the class names \cite{parashar2024neglected} and then finetunes both the visual encoder and classifier head using few-shot labeled data. 
It adopts a learning rate of 1e-4 for the classifier and a smaller learning rate of 1e-6 for the visual encoder to preserve the pretrained features \cite{wortsman2022robust}.
It uses a batch size of 32, weight decay of 1e-2, \emph{AdamW} optimizer, and a cosine annealing learning rate scheduler to train the model for 20 epochs.
For the DINOv2 ViT-B/14 backbone, we adopt the same hyperparameters except for initializing the classifier by linear probing on the few-shot data for 50 epochs.
For models pretrained on ImageNet \cite{deng2009imagenet} or iNaturalist \cite{inat2021}, we adopt the hyperparameters from \cite{su2021realistic}, using a learning rate of 1e-3 for both backbone and classifier. Similarly, it initializes the classifier by linear probing on few-shot labeled data.

{\bf Other FSL methods.}
We run other FSL methods using the OpenCLIP ViT-B/32 \cite{cherti2023reproducible} model with their reported hyperparameters for all datasets.

\section{Code and Instructions}
\label{sec:Demo-code}

We release open-source Python code 
at \url{https://tian1327.github.io/POC}.
We provide usage instructions below.

{\bf Dependencies}.
Running our code requires packages, such as \emph{torchvision}, \emph{PyTorch},
\emph{clip}, and \emph{open\_clip\_torch}.
We provide detailed instructions for building the environment in file {\tt README.md}.
We suggest assigning $>$50GB storage space and $>$40GB GPU RAM to reproduce our experiments.

{\bf License}.
We release code under the MIT License to foster future research.

{\bf Instructions.}
We provide detailed step-by-step instructions for running our code in markdown files.

\begin{itemize}

\item
\begin{verbatim}README.md\end{verbatim}
We provide instructions to set up the environment and run various FSL baselines, as well as POC with different LMMs. In addition, we provide guidelines to reproduce our results with FineR \cite{liu2024democratizing}.

\item
\begin{verbatim}DATASETS.md\end{verbatim}
We provide detailed steps for setting up the benchmarking datasets and sampling few-shot data from the official training sets.

\end{itemize}

\section{Prompt Templates}
\label{sec:prompt_templates}

We provide examples of different prompting strategies in \cref{fig:pt_templates}, including
open-vocabulary prompting,
open-vocabulary with Chain-of-Thought \cite{kojima2022large} or self-verification \cite{weng2023large}, In-Context Learning (ICL) with all class names \cite{jiang2405many}.
Finally, \cref{fig:mmICL-prompt} of the main paper presents our Post-Hoc Correction (POC) prompt, which incorporates the expert model’s top-$k$ predicted class names, their confidence scores, and the stitched few-shot visual exemplars, followed by a re-ranking instruction. 
Such multimodal prompts achieve the best performance on five VSR benchmarks, yielding significant gains over the few-shot expert model.
We postpone the ablation study until \cref{tab:abl_POC_v2}.

\section{Detailed Benchmarking Results}
\label{sec:detailed_results}

We provide detailed per-dataset results for experiments shown in the main paper, along with additional analyses below.

\begin{figure}[t] 
\centering
\includegraphics[width=\linewidth]{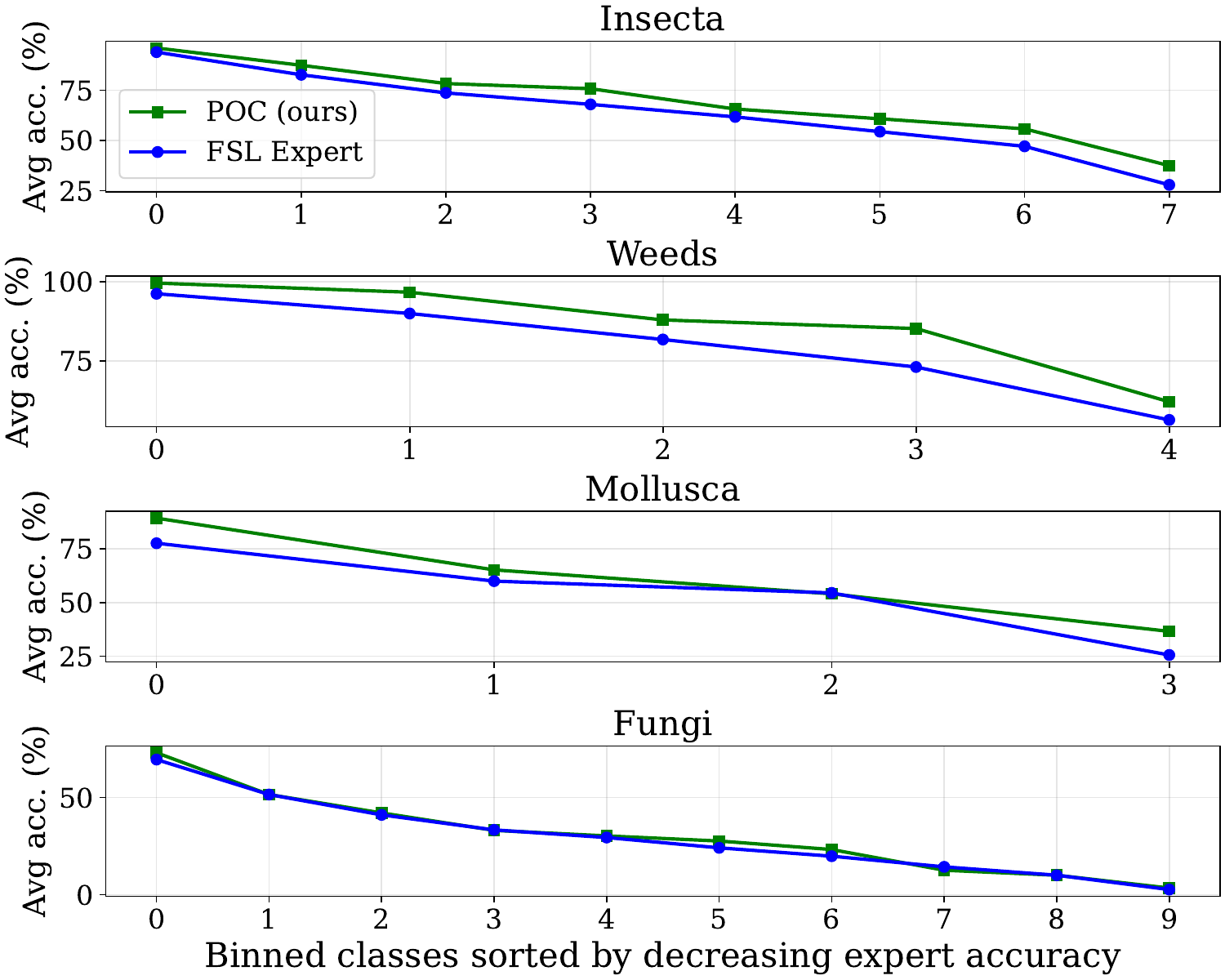}
\vspace{-6mm}
\caption{
\small
\textbf{Improvement of POC over FSL expert model on binned classes for each benchmark.} 
Using the same expert and LMM as \cref{tab:abl_POC_v2},
POC improves on both easy and hard classes.} \label{fig:improve_acc_more}
\end{figure}

\textbf{Analysis of POC's improvement over the expert model.}
\cref{fig:improve_acc_more} supplements \cref{fig:improve_acc} in the main paper by providing results on other benchmarks.
Results show that POC generally improves on easy and hard classes.

\begin{table*}[t]
\centering
\small
\caption{\small
{\bf POC significantly improves existing FSL methods across five benchmarks under various few-shot settings.}
We run POC with the Qwen-2.5-VL-7B LMM \cite{qwen2.5-vl} and expert models learned by different FSL methods atop the OpenCLIP ViT-B/32 backbone \cite{cherti2023reproducible},
under the 4-, 8-, and 16-shot settings.
The FSL methods include prompt learning (PL) \cite{zhou2022learning, maple}, adapter learning (AL) \cite{clipadapter, tipadapter}, linear probing (LP) \cite{radford2021learning, lin2023multimodality, clap24}, full finetuning (FT) \cite{liu2025few}, and the non-learned (NL) FineR \cite{liu2024democratizing}. \textcolor{Green}{Superscripts} denote the accuracy gains of POC over the corresponding FSL method.
Results show that POC consistently improves all FSL methods across various benchmarks and few-shot settings, with larger gains in lower-shot settings.
{\bf Bold} and \underline{underlined} numbers mark the best and second-best results under each shot setting.
}
\label{tab:improve_fsr_detail}

\vspace{-2mm}
\setlength{\tabcolsep}{0.35em}
\scalebox{0.9}{
\begin{tabular}{cllcccccccccccc}
\toprule
\multirow{2}{*}{shots} &
\multirow{2}{*}{\makecell{}} &
\multirow{2}{*}{FSL method} &
\multicolumn{2}{c}{Aves} &
\multicolumn{2}{c}{Insecta} &
\multicolumn{2}{c}{Weeds} &
\multicolumn{2}{c}{Mollusca} &
\multicolumn{2}{c}{Fungi} &
\multicolumn{2}{c}{\cellcolor{col33}mean acc.} \\

\cmidrule(lr){4-5}\cmidrule(lr){6-7}\cmidrule(lr){8-9}\cmidrule(lr){10-11}\cmidrule(lr){12-13}\cmidrule(lr){14-15}
& & &
expert & POC &
expert & POC &
expert & POC &
expert & POC &
expert & POC &
\cellcolor{col33}expert & \cellcolor{col33}POC \\
\midrule
\multirow{11}{*}{4}
& \multirow{2}{*}{\rotatebox[origin=c]{0}{\makecell{PL}}}
  & CoOp \cite{zhou2022learning}\textsubscript{\textcolor{gray}{IJCV'22}}  &
  42.6 & 58.0$^{\textcolor{Green}{+15.4}}$ &
  14.7 & 24.9$^{\textcolor{Green}{+10.2}}$ &
  51.4 & 73.1$^{\textcolor{Green}{+21.7}}$ &
  41.1 & 52.8$^{\textcolor{Green}{+11.7}}$ &
   3.2 &  5.3$^{\textcolor{Green}{+2.1}}$  &
\cellcolor{col33}30.6 & \cellcolor{col33}42.8$^{\textcolor{Green}{+12.2}}$ \\

& & MaPLe \cite{maple}\textsubscript{\textcolor{gray}{CVPR'23}}  &
  44.2 & 58.6$^{\textcolor{Green}{+14.4}}$ &
   9.8 & 19.0$^{\textcolor{Green}{+9.2}}$  &
  40.7 & 58.7$^{\textcolor{Green}{+18.0}}$ &
  22.7 & 42.6$^{\textcolor{Green}{+19.9}}$ &
   1.5 &  3.2$^{\textcolor{Green}{+1.7}}$  &
\cellcolor{col33}23.8 & \cellcolor{col33}36.4$^{\textcolor{Green}{+12.6}}$ \\
\cmidrule{2-15}

& \multirow{2}{*}{\rotatebox[origin=c]{0}{\makecell{AL}}}
  & Tip-Adapter \cite{tipadapter}\textsubscript{\textcolor{gray}{ECCV'22}}  &
  45.9 & 58.9$^{\textcolor{Green}{+13.0}}$ &
  11.4 & 20.5$^{\textcolor{Green}{+9.1}}$  &
  48.6 & 64.5$^{\textcolor{Green}{+15.9}}$ &
  19.9 & 41.1$^{\textcolor{Green}{+21.2}}$ &
   1.7 &  3.1$^{\textcolor{Green}{+1.4}}$  &
\cellcolor{col33}25.5 & \cellcolor{col33}37.6$^{\textcolor{Green}{+12.1}}$ \\

& & CLIP-Adapter \cite{clipadapter}\textsubscript{\textcolor{gray}{IJCV'23}}  &
  46.0 & 60.3$^{\textcolor{Green}{+14.3}}$ &
  16.1 & 26.6$^{\textcolor{Green}{+10.5}}$ &
  50.6 & 72.5$^{\textcolor{Green}{+21.9}}$ &
  37.8 & 50.6$^{\textcolor{Green}{+12.8}}$ &
   2.5 &  4.5$^{\textcolor{Green}{+2.0}}$  &
\cellcolor{col33}30.6 & \cellcolor{col33}42.9$^{\textcolor{Green}{+12.3}}$ \\
\cmidrule{2-15}

& \multirow{3}{*}{\rotatebox[origin=c]{0}{\makecell{LP}}}
  & Few-shot LP \cite{radford2021learning}\textsubscript{\textcolor{gray}{ICML'21}}  &
  45.0 & 61.2$^{\textcolor{Green}{+16.2}}$ &
  26.0 & 39.7$^{\textcolor{Green}{+13.7}}$ &
  54.0 & 77.1$^{\textcolor{Green}{+23.1}}$ &
  37.9 & 52.8$^{\textcolor{Green}{+14.9}}$ &
   9.0 & 11.2$^{\textcolor{Green}{+2.2}}$  &
\cellcolor{col33}34.4 & \cellcolor{col33}48.4$^{\textcolor{Green}{+14.0}}$ \\

& & CrossModal LP \cite{lin2023multimodality}\textsubscript{\textcolor{gray}{CVPR'23}}  &
  \underline{47.4} & 61.1$^{\textcolor{Green}{+13.7}}$ &
  30.1 & 46.1$^{\textcolor{Green}{+16.0}}$ &
  58.2 & 78.2$^{\textcolor{Green}{+20.0}}$ &
  37.1 & 50.7$^{\textcolor{Green}{+13.6}}$ &
   9.3 & 12.2$^{\textcolor{Green}{+2.9}}$  &
\cellcolor{col33}36.4 & \cellcolor{col33}49.7$^{\textcolor{Green}{+13.2}}$ \\

& & CLAP \cite{clap24}\textsubscript{\textcolor{gray}{CVPR'24}}  &
  \textbf{49.9} & \underline{62.3}$^{\textcolor{Green}{+12.4}}$ &
  \textbf{39.3} & \textbf{53.0}$^{\textcolor{Green}{+13.7}}$ &
  \textbf{68.0} & \textbf{83.5}$^{\textcolor{Green}{+15.5}}$ &
  \textbf{53.6} & \textbf{61.8}$^{\textcolor{Green}{+8.2}}$  &
  \textbf{12.3} & \textbf{15.0}$^{\textcolor{Green}{+2.7}}$  &
\cellcolor{col33}\textbf{44.6} & \cellcolor{col33}\textbf{55.1}$^{\textcolor{Green}{+10.5}}$ \\
\cmidrule{2-15}

& \multirow{1}{*}{\rotatebox[origin=c]{0}{NL}}
  & FineR \cite{liu2024democratizing}\textsubscript{\textcolor{gray}{ICLR'24}}  &
  42.9 & 58.7$^{\textcolor{Green}{+15.8}}$ &
  27.7 & 42.0$^{\textcolor{Green}{+14.3}}$ &
  \underline{61.2} & \underline{81.4}$^{\textcolor{Green}{+20.2}}$ &
  41.8 & 54.0$^{\textcolor{Green}{+12.2}}$ &
   7.2 &  9.7$^{\textcolor{Green}{+2.5}}$  &
\cellcolor{col33}36.2 & \cellcolor{col33}49.2$^{\textcolor{Green}{+13.0}}$ \\
\cmidrule{2-15}

& \multirow{1}{*}{\rotatebox[origin=c]{0}{FT}}
  & Few-shot FT \cite{liu2025few}\textsubscript{\textcolor{gray}{CVPR'25}}  &
  46.8 & \textbf{63.0}$^{\textcolor{Green}{+16.2}}$ &
  \underline{36.5} & \underline{48.5}$^{\textcolor{Green}{+12.0}}$ &
  56.2 & 79.5$^{\textcolor{Green}{+23.3}}$ &
  \underline{44.6} & \underline{55.5}$^{\textcolor{Green}{+10.9}}$ &
  \underline{12.1} & \underline{13.6}$^{\textcolor{Green}{+1.5}}$  &
\cellcolor{col33}\underline{39.2} & \cellcolor{col33}\underline{52.0}$^{\textcolor{Green}{+12.8}}$ \\

\midrule
\multirow{11}{*}{8}
& \multirow{2}{*}{\rotatebox[origin=c]{0}{\makecell{PL}}}
  & CoOp \cite{zhou2022learning}\textsubscript{\textcolor{gray}{IJCV'22}}  &
  45.6 & 61.6$^{\textcolor{Green}{+16.0}}$ &
  26.4 & 42.8$^{\textcolor{Green}{+16.4}}$ &
  57.9 & 77.6$^{\textcolor{Green}{+19.7}}$ &
  36.1 & 54.6$^{\textcolor{Green}{+18.5}}$ &
   5.3 &  8.8$^{\textcolor{Green}{+3.5}}$  &
\cellcolor{col33}34.3 & \cellcolor{col33}49.1$^{\textcolor{Green}{+14.8}}$ \\

& & MaPLe \cite{maple}\textsubscript{\textcolor{gray}{CVPR'23}}  &
  46.4 & 60.4$^{\textcolor{Green}{+14.0}}$ &
   9.8 & 22.2$^{\textcolor{Green}{+12.4}}$ &
  41.0 & 59.3$^{\textcolor{Green}{+18.3}}$ &
  23.2 & 43.2$^{\textcolor{Green}{+20.0}}$ &
   2.5 &  5.2$^{\textcolor{Green}{+2.7}}$  &
\cellcolor{col33}24.6 & \cellcolor{col33}38.1$^{\textcolor{Green}{+13.5}}$ \\
\cmidrule{2-15}

& \multirow{2}{*}{\rotatebox[origin=c]{0}{\makecell{AL}}}
  & Tip-Adapter \cite{tipadapter}\textsubscript{\textcolor{gray}{ECCV'22}}  &
  46.4 & 59.7$^{\textcolor{Green}{+13.3}}$ &
  12.3 & 23.0$^{\textcolor{Green}{+10.7}}$ &
  50.2 & 65.6$^{\textcolor{Green}{+15.4}}$ &
  20.9 & 41.1$^{\textcolor{Green}{+20.2}}$ &
   2.2 &  5.1$^{\textcolor{Green}{+2.9}}$  &
\cellcolor{col33}26.4 & \cellcolor{col33}38.9$^{\textcolor{Green}{+12.5}}$ \\

& & CLIP-Adapter \cite{clipadapter}\textsubscript{\textcolor{gray}{IJCV'23}}  &
  47.7 & 62.6$^{\textcolor{Green}{+14.9}}$ &
  22.3 & 40.1$^{\textcolor{Green}{+17.8}}$ &
  54.2 & 76.2$^{\textcolor{Green}{+22.0}}$ &
  37.6 & 51.7$^{\textcolor{Green}{+14.1}}$ &
   3.9 &  7.7$^{\textcolor{Green}{+3.8}}$  &
\cellcolor{col33}33.1 & \cellcolor{col33}47.7$^{\textcolor{Green}{+14.5}}$ \\
\cmidrule{2-15}

& \multirow{3}{*}{\rotatebox[origin=c]{0}{\makecell{LP}}}
  & Few-shot LP \cite{radford2021learning}\textsubscript{\textcolor{gray}{ICML'21}}  &
  49.8 & 64.9$^{\textcolor{Green}{+15.1}}$ &
  41.6 & 54.7$^{\textcolor{Green}{+13.1}}$ &
  59.3 & 79.6$^{\textcolor{Green}{+20.3}}$ &
  32.4 & 54.7$^{\textcolor{Green}{+22.3}}$ &
  13.2 & 16.5$^{\textcolor{Green}{+3.3}}$  &
\cellcolor{col33}39.3 & \cellcolor{col33}54.1$^{\textcolor{Green}{+14.8}}$ \\

& & CrossModal LP \cite{lin2023multimodality}\textsubscript{\textcolor{gray}{CVPR'23}}  &
  50.8 & 64.1$^{\textcolor{Green}{+13.3}}$ &
  45.6 & 58.8$^{\textcolor{Green}{+13.2}}$ &
  65.4 & 80.1$^{\textcolor{Green}{+14.7}}$ &
  \underline{47.2} & \underline{57.0}$^{\textcolor{Green}{+9.8}}$  &
  13.8 & 18.2$^{\textcolor{Green}{+4.4}}$  &
\cellcolor{col33}44.6 & \cellcolor{col33}55.6$^{\textcolor{Green}{+11.1}}$ \\

& & CLAP \cite{clap24}\textsubscript{\textcolor{gray}{CVPR'24}}  &
  \textbf{53.6} & \underline{65.9}$^{\textcolor{Green}{+12.3}}$ &
  \textbf{55.2} & \textbf{65.3}$^{\textcolor{Green}{+10.1}}$ &
  \underline{73.9} & 80.5$^{\textcolor{Green}{+6.6}}$  &
  \textbf{59.5} & \textbf{64.8}$^{\textcolor{Green}{+5.3}}$  &
  \underline{18.4} & \underline{22.5}$^{\textcolor{Green}{+4.1}}$  &
\cellcolor{col33}\textbf{52.1} & \cellcolor{col33}\textbf{59.8}$^{\textcolor{Green}{+7.7}}$ \\
\cmidrule{2-15}

& \multirow{1}{*}{\rotatebox[origin=c]{0}{NL}}
  & FineR \cite{liu2024democratizing}\textsubscript{\textcolor{gray}{ICLR'24}}  &
  46.6 & 61.1$^{\textcolor{Green}{+14.5}}$ &
  32.9 & 46.8$^{\textcolor{Green}{+13.9}}$ &
  \textbf{74.3} & \underline{81.2}$^{\textcolor{Green}{+6.9}}$  &
  38.6 & 54.6$^{\textcolor{Green}{+17.5}}$ &
   8.6 & 12.0$^{\textcolor{Green}{+3.4}}$  &
\cellcolor{col33}39.9 & \cellcolor{col33}51.1$^{\textcolor{Green}{+11.2}}$ \\
\cmidrule{2-15}

& \multirow{1}{*}{\rotatebox[origin=c]{0}{FT}}
  & Few-shot FT \cite{liu2025few}\textsubscript{\textcolor{gray}{CVPR'25}}  &
  \underline{53.0} & \textbf{66.1}$^{\textcolor{Green}{+13.1}}$ &
  \underline{54.4} & \underline{63.4}$^{\textcolor{Green}{+9.0}}$  &
  70.4 & \textbf{83.8}$^{\textcolor{Green}{+13.4}}$ &
  39.4 & 56.4$^{\textcolor{Green}{+17.0}}$ &
  \textbf{20.1} & \textbf{22.9}$^{\textcolor{Green}{+2.8}}$  &
\cellcolor{col33}\underline{47.4} & \cellcolor{col33}\underline{58.5}$^{\textcolor{Green}{+11.1}}$ \\

\midrule
\multirow{11}{*}{16}
& \multirow{2}{*}{\rotatebox[origin=c]{0}{\makecell{PL}}}
  & CoOp \cite{zhou2022learning}\textsubscript{\textcolor{gray}{IJCV'22}}  &
  48.2 & 62.0$^{\textcolor{Green}{+13.8}}$ &
  38.2 & 51.4$^{\textcolor{Green}{+13.2}}$ &
  69.9 & 83.7$^{\textcolor{Green}{+13.8}}$ &
  44.5 & 59.7$^{\textcolor{Green}{+15.2}}$ &
   8.9 & 13.0$^{\textcolor{Green}{+4.1}}$ &
\cellcolor{col33}41.9 & \cellcolor{col33}53.9$^{\textcolor{Green}{+12.0}}$ \\

& & MaPLe \cite{maple}\textsubscript{\textcolor{gray}{CVPR'23}}  &
  47.4 & 61.0$^{\textcolor{Green}{+13.6}}$ &
  15.7 & 29.2$^{\textcolor{Green}{+13.5}}$ &
  41.5 & 60.6$^{\textcolor{Green}{+19.1}}$ &
  23.6 & 42.9$^{\textcolor{Green}{+19.3}}$ &
   3.1 &  6.6$^{\textcolor{Green}{+3.5}}$ &
\cellcolor{col33}26.3 & \cellcolor{col33}40.0$^{\textcolor{Green}{+13.7}}$ \\
\cmidrule{2-15}

& \multirow{2}{*}{\rotatebox[origin=c]{0}{\makecell{AL}}}
  & Tip-Adapter \cite{tipadapter}\textsubscript{\textcolor{gray}{ECCV'22}}  &
  47.7 & 59.6$^{\textcolor{Green}{+11.9}}$ &
  14.6 & 28.3$^{\textcolor{Green}{+13.7}}$ &
  53.1 & 71.7$^{\textcolor{Green}{+18.6}}$ &
  27.3 & 44.7$^{\textcolor{Green}{+17.4}}$ &
   3.2 &  7.4$^{\textcolor{Green}{+4.2}}$ &
\cellcolor{col33}29.2 & \cellcolor{col33}42.3$^{\textcolor{Green}{+13.1}}$ \\

& & CLIP-Adapter \cite{clipadapter}\textsubscript{\textcolor{gray}{IJCV'23}}  &
  50.9 & 63.6$^{\textcolor{Green}{+12.7}}$ &
  38.8 & 56.3$^{\textcolor{Green}{+17.5}}$ &
  68.4 & 82.0$^{\textcolor{Green}{+13.6}}$ &
  48.4 & 57.4$^{\textcolor{Green}{+9.0}}$ &
   6.3 & 10.8$^{\textcolor{Green}{+4.5}}$ &
\cellcolor{col33}42.6 & \cellcolor{col33}54.0$^{\textcolor{Green}{+11.4}}$ \\
\cmidrule{2-15}

& \multirow{3}{*}{\rotatebox[origin=c]{0}{\makecell{LP}}}
  & Few-shot LP \cite{radford2021learning}\textsubscript{\textcolor{gray}{ICML'21}}  &
  53.8 & 67.7$^{\textcolor{Green}{+13.9}}$ &
  54.7 & 63.0$^{\textcolor{Green}{+8.3}}$ &
  69.7 & 85.2$^{\textcolor{Green}{+15.5}}$ &
  56.8 & 63.9$^{\textcolor{Green}{+7.1}}$ &
  20.0 & 22.7$^{\textcolor{Green}{+2.7}}$ &
\cellcolor{col33}51.0 & \cellcolor{col33}60.5$^{\textcolor{Green}{+9.5}}$ \\

& & CrossModal LP \cite{lin2023multimodality}\textsubscript{\textcolor{gray}{CVPR'23}}  &
  53.3 & 66.0$^{\textcolor{Green}{+12.7}}$ &
  57.3 & 66.6$^{\textcolor{Green}{+9.3}}$ &
  75.1 & 84.3$^{\textcolor{Green}{+9.2}}$ &
  58.2 & 65.2$^{\textcolor{Green}{+7.0}}$ &
  21.8 & 25.3$^{\textcolor{Green}{+3.5}}$ &
\cellcolor{col33}53.1 & \cellcolor{col33}61.5$^{\textcolor{Green}{+8.4}}$ \\

& & CLAP \cite{clap24}\textsubscript{\textcolor{gray}{CVPR'24}}  &
  \underline{57.0} & \underline{67.8}$^{\textcolor{Green}{+10.8}}$ &
  \underline{63.1} & \textbf{70.8}$^{\textcolor{Green}{+7.7}}$ &
  \underline{76.9} & \underline{85.0}$^{\textcolor{Green}{+8.1}}$ &
  \underline{63.5} & \textbf{71.6}$^{\textcolor{Green}{+8.1}}$ &
  \underline{24.9} & \underline{28.2}$^{\textcolor{Green}{+3.3}}$ &
\cellcolor{col33}\underline{57.1} & \cellcolor{col33}\underline{64.7}$^{\textcolor{Green}{+7.6}}$ \\
\cmidrule{2-15}

& \multirow{1}{*}{\rotatebox[origin=c]{0}{NL}}
  & FineR \cite{liu2024democratizing}\textsubscript{\textcolor{gray}{ICLR'24}}  &
  47.5 & 62.5$^{\textcolor{Green}{+15.0}}$ &
  32.8 & 47.1$^{\textcolor{Green}{+14.3}}$ &
  65.0 & 82.4$^{\textcolor{Green}{+17.4}}$ &
  47.7 & 58.7$^{\textcolor{Green}{+11.0}}$ &
   8.9 & 13.9$^{\textcolor{Green}{+5.0}}$ &
\cellcolor{col33}40.4 & \cellcolor{col33}52.9$^{\textcolor{Green}{+12.5}}$ \\
\cmidrule{2-15}

& \multirow{1}{*}{\rotatebox[origin=c]{0}{FT}}
  & Few-shot FT \cite{liu2025few}\textsubscript{\textcolor{gray}{CVPR'25}}  &
  \textbf{58.2} & \textbf{69.4}$^{\textcolor{Green}{+11.2}}$ &
  \textbf{63.8} & \underline{70.5}$^{\textcolor{Green}{+6.7}}$ &
  \textbf{80.7} & \textbf{87.7}$^{\textcolor{Green}{+7.0}}$ &
  \textbf{63.6} & \underline{69.2}$^{\textcolor{Green}{+5.6}}$ &
  \textbf{29.9} & \textbf{31.1}$^{\textcolor{Green}{+1.2}}$ &
\cellcolor{col33}\textbf{59.2} & \cellcolor{col33}\textbf{65.6}$^{\textcolor{Green}{+6.4}}$ \\
\bottomrule
\end{tabular}}
\end{table*}

{
\setlength{\tabcolsep}{1.5em}
\newcommand{\bodystrut}{\rule{0pt}{2.6ex}}

\begin{table*}[t]
\centering
\caption{\small \textbf{Per-dataset comparison of POC using an expert model trained with different pretrained backbones}. 
We train an expert model for each backbone by finetuning a pretrained visual encoder using 16-shot labeled data \cite{liu2025few} randomly sampled from each dataset with three random seeds.
We report the mean accuracy and standard deviation in subscripts. 
Results show that POC yields consistent gains over the expert model trained with different backbones, with larger gains on weaker encoders such as ImageNet-pretrained ResNet-50 \cite{deng2009imagenet, he2016deep}. The results highlight the advantages of POC for serving as a simple plug-and-play module.
\textcolor{Green}{Superscripts} denote the accuracy gains over the expert model. 
}
\label{tab:backbone_detailed}
\vspace{-2mm}

\scalebox{0.85}{
\begin{tabular}{lcccccccc}
\toprule

 &
\multicolumn{2}{c}{ImageNet-RN50 \cite{deng2009imagenet}} &
\multicolumn{2}{c}{OpenCLIP ViT-B/32 \cite{cherti2023reproducible}} &
\multicolumn{2}{c}{iNaturalist-RN50 \cite{inat2017}} &
\multicolumn{2}{c}{DINOv2 ViT-B/14 \cite{oquab2023dinov2}} \\

\cmidrule(lr){2-3}
\cmidrule(lr){4-5}
\cmidrule(lr){6-7}
\cmidrule(lr){8-9}

dataset &
expert & POC &
expert & POC &
expert & POC &
expert & POC \\

\midrule

Aves\bodystrut &
$44.3_{\,\pm0.6}$ & $56.4_{\,\pm0.3}^{\textcolor{Green}{+12.1}}$ &
$58.3_{\,\pm0.2}$ & $69.3_{\,\pm0.7}^{\textcolor{Green}{+11.0}}$ &
$63.6_{\,\pm0.4}$ & $70.9_{\,\pm0.5}^{\textcolor{Green}{+7.3}}$ &
$79.3_{\,\pm0.3}$ & $82.2_{\,\pm0.4}^{\textcolor{Green}{+2.9}}$ \\

Insecta\bodystrut &
$64.1_{\,\pm1.4}$ & $71.1_{\,\pm1.3}^{\textcolor{Green}{+7.0}}$ &
$65.7_{\,\pm1.9}$ & $72.0_{\,\pm1.5}^{\textcolor{Green}{+6.3}}$ &
$69.9_{\,\pm0.2}$ & $74.8_{\,\pm0.2}^{\textcolor{Green}{+4.9}}$ &
$77.5_{\,\pm1.3}$ & $80.0_{\,\pm0.8}^{\textcolor{Green}{+2.5}}$ \\

Weeds\bodystrut &
$73.0_{\,\pm0.3}$ & $84.2_{\,\pm0.5}^{\textcolor{Green}{+11.2}}$ &
$80.7_{\,\pm0.6}$ & $87.9_{\,\pm0.2}^{\textcolor{Green}{+7.2}}$ &
$84.4_{\,\pm0.7}$ & $90.2_{\,\pm0.7}^{\textcolor{Green}{+5.8}}$ &
$93.3_{\,\pm0.1}$ & $94.8_{\,\pm0.3}^{\textcolor{Green}{+1.5}}$ \\

Mollusca\bodystrut &
$60.4_{\,\pm3.0}$ & $69.2_{\,\pm2.7}^{\textcolor{Green}{+8.8}}$ &
$63.3_{\,\pm0.4}$ & $68.3_{\,\pm0.9}^{\textcolor{Green}{+5.0}}$ &
$63.6_{\,\pm2.5}$ & $68.5_{\,\pm2.8}^{\textcolor{Green}{+4.9}}$ &
$49.4_{\,\pm1.6}$ & $65.1_{\,\pm0.8}^{\textcolor{Green}{+15.7}}$ \\

Fungi\bodystrut &
$25.4_{\,\pm0.8}$ & $26.9_{\,\pm1.1}^{\textcolor{Green}{+1.5}}$ &
$30.1_{\,\pm0.4}$ & $31.5_{\,\pm0.4}^{\textcolor{Green}{+1.4}}$ &
$26.3_{\,\pm0.8}$ & $28.0_{\,\pm0.7}^{\textcolor{Green}{+1.7}}$ &
$41.1_{\,\pm0.5}$ & $41.4_{\,\pm0.4}^{\textcolor{Green}{+0.3}}$ \\

\rowcolor{col33}
mean acc.\bodystrut &
$53.5_{\,\pm0.7}$ & \textbf{$61.5_{\,\pm0.7}^{\textcolor{Green}{+8.0}}$} &
$59.6_{\,\pm0.5}$ & \textbf{$65.8_{\,\pm0.3}^{\textcolor{Green}{+6.2}}$} &
$61.5_{\,\pm0.7}$ & \textbf{$66.5_{\,\pm0.8}^{\textcolor{Green}{+5.0}}$} &
$68.1_{\,\pm0.1}$ & \textbf{$72.7_{\,\pm0.3}^{\textcolor{Green}{+4.6}}$} \\

\bottomrule
\end{tabular}}
\end{table*}
}

\textbf{POC improves various existing FSL methods.}
\cref{tab:improve_fsr_detail} supplements \cref{tab:improve_fsr} in the main paper by validating that POC consistently improves various FSL methods across 4-shot and 8-shot settings.
Importantly, the accuracy gains are larger in lower-shot settings, highlighting the advantages of POC for challenging data-scarce scenarios.

\textbf{POC generalizes to different pretrained backbones.}
\cref{tab:backbone_detailed} reports per-dataset results corresponding to \cref{fig:ablate_backbone}, using expert models trained on different pretrained backbones. POC consistently improves performance across all backbones.

\textbf{POC generalizes to different LMMs.}
\cref{tab:lmm_poc_summary} provides the per-dataset results of POC with different LMMs. The consistent improvement across datasets validates that POC generalizes to different LMMs, serving as a simple plug-and-play module to improve few-shot methods on VSR tasks.

\textbf{POC also improves non-VSR tasks.} 
We experiment with POC on popular non-VSR fine-grained benchmarks (\cref{tab:datasets}), which are widely used in FSL literature \cite{liu2025few, clap24, lin2023multimodality, zhou2022learning, radford2021learning}. 
Following the same experiment setup, we randomly sample $K$-shot ($K=16$) labeled data and use the same backbone (OpenCLIP ViT-B/32) and LMM (Qwen-2.5-VL-7B-Instruct). As shown in \cref{tab:nonvsr_generalization}, POC consistently improves over the FSL expert across all benchmarks, demonstrating the practical value of POC beyond species recognition.

{
\setlength{\tabcolsep}{0.25em}  
\begin{table}[ht]
\centering
\footnotesize
\caption{\small
{\bf
Comparison between POC and different prompting strategies using various LMMs.}
As baselines, we list the zero-shot accuracy of OpenCLIP ViT-B/32 \cite{cherti2023reproducible} and the performance of FSL expert that finetunes OpenCLIP on 16-shot labeled data for each dataset \cite{liu2025few}.
We compared various prompting strategies (templates shown in \cref{fig:pt_templates}) with POC, using GLM-4.1V-9B-Thinking, Qwen-2.5-VL-7B, and the closed-source GPT-5-Mini.
Results show that all LMMs significantly underperform the FSL expert (cf. \textcolor{Red}{red superscripts}), even with established Chain-of-Thought prompting \cite{kojima2022large} or In-Context Learning \cite{jiang2405many}.
In contrast, our POC's multimodal prompt yields significant accuracy gains over the expert model (cf. \textcolor{Green}{green superscripts}).
}
\vspace{-2mm}
\label{tab:lmm_poc_summary}
\scalebox{0.85}{
\begin{tabular}{lllllll}
\toprule
 model & Aves & Insecta & Weeds & Mollusca & Fungi & \cellcolor{col33}mean acc. \\
\midrule
\rowcolor{gray!15}
OpenCLIP \cite{cherti2023reproducible}
         & 44.8 & 10.0 & 47.2 & 19.7 & 2.1 & \cellcolor{col33}24.8 \\
\rowcolor{gray!15}
FSL Expert \cite{liu2025few}
         & 58.2 & 63.8 & 80.7 & 63.6 & 29.9 & \cellcolor{col33}59.2 \\
\midrule

\multicolumn{7}{l}{\emph{Open Vocabulary Prompting}} \\
GLM-4.1V-9B   & 35.4$^{\textcolor{Red}{-22.8}}$
         &  7.0$^{\textcolor{Red}{-56.8}}$
         & 12.7$^{\textcolor{Red}{-68.0}}$
         & 27.1$^{\textcolor{Red}{-36.5}}$
         &  4.0$^{\textcolor{Red}{-25.9}}$
         & \cellcolor{col33}17.2$^{\textcolor{Red}{-42.0}}$ \\
Qwen-2.5-VL-7B   & 37.7$^{\textcolor{Red}{-20.5}}$
         & 13.5$^{\textcolor{Red}{-50.3}}$
         & 17.0$^{\textcolor{Red}{-63.7}}$
         & 35.9$^{\textcolor{Red}{-27.7}}$
         &  3.0$^{\textcolor{Red}{-26.9}}$
         & \cellcolor{col33}21.4$^{\textcolor{Red}{-37.8}}$ \\
GPT-5-Mini    & 32.8$^{\textcolor{Red}{-25.4}}$
         &  9.5$^{\textcolor{Red}{-54.3}}$
         &  9.6$^{\textcolor{Red}{-71.1}}$
         & 17.3$^{\textcolor{Red}{-46.3}}$
         &  5.0$^{\textcolor{Red}{-24.9}}$
         & \cellcolor{col33}14.8$^{\textcolor{Red}{-44.4}}$ \\

\midrule

\multicolumn{7}{l}{\emph{Chain-of-Thought Prompting}} \\
Qwen-2.5-VL-7B
& 34.2$^{\textcolor{Red}{-24.0}}$ 
& 10.9$^{\textcolor{Red}{-52.9}}$ 
& 14.6$^{\textcolor{Red}{-66.1}}$ 
& 32.0$^{\textcolor{Red}{-31.6}}$ 
& 3.4$^{\textcolor{Red}{-26.5}}$ 
& \cellcolor{col33}19.0$^{\textcolor{Red}{-40.2}}$ \\

GPT-5-Mini
& 20.4$^{\textcolor{Red}{-37.8}}$ 
& 5.8$^{\textcolor{Red}{-58.0}}$
& 5.5$^{\textcolor{Red}{-75.2}}$ 
& 12.5$^{\textcolor{Red}{-51.1}}$ 
& 1.2$^{\textcolor{Red}{-28.7}}$ 
& \cellcolor{col33}9.1$^{\textcolor{Red}{-50.1}}$ \\

\midrule

\multicolumn{7}{l}{\emph{In-Context Learning Prompting (w/ all class names)}} \\
Qwen-2.5-VL-7B
& 35.9$^{\textcolor{Red}{-22.3}}$ 
& 17.3$^{\textcolor{Red}{-46.5}}$ 
& 78.1$^{\textcolor{Red}{-2.6}}$ 
& 45.2$^{\textcolor{Red}{-18.4}}$ 
& 1.4$^{\textcolor{Red}{-28.5}}$ 
& \cellcolor{col33}35.6$^{\textcolor{Red}{-23.6}}$ \\

GPT-5-Mini
& 54.5$^{\textcolor{Red}{-3.7}}$ 
& 33.6$^{\textcolor{Red}{-30.2}}$ 
& 79.2$^{\textcolor{Red}{-1.5}}$ 
& 44.7$^{\textcolor{Red}{-18.9}}$ 
& 6.7$^{\textcolor{Red}{-23.2}}$ 
& \cellcolor{col33}43.7$^{\textcolor{Red}{-15.5}}$ \\

\midrule
\multicolumn{7}{l}{\emph{\textbf{POC prompting (ours)}}} \\
GLM-4.1V-9B  & 65.0$^{\textcolor{Green}{+6.8}}$
         & 68.3$^{\textcolor{Green}{+4.5}}$
         & 85.7$^{\textcolor{Green}{+5.0}}$
         & 66.3$^{\textcolor{Green}{+2.7}}$
         & \underline{31.8}$^{\textcolor{Green}{+1.9}}$
         & \cellcolor{col33}63.4$^{\textcolor{Green}{+4.2}}$ \\
Qwen-2.5-VL-7B & \textbf{69.4}$^{\textcolor{Green}{+11.2}}$
         & \underline{70.5}$^{\textcolor{Green}{+6.7}}$
         & \textbf{87.7}$^{\textcolor{Green}{+7.0}}$
         & \underline{69.2}$^{\textcolor{Green}{+5.6}}$
         & 31.1$^{\textcolor{Green}{+1.2}}$
         & \cellcolor{col33}\underline{65.6}$^{\textcolor{Green}{+6.4}}$ \\
GPT-5-Mini  & \underline{66.8}$^{\textcolor{Green}{+8.6}}$
         & \textbf{70.6}$^{\textcolor{Green}{+6.8}}$
         & \underline{87.3}$^{\textcolor{Green}{+6.6}}$
         & \textbf{71.6}$^{\textcolor{Green}{+8.0}}$
         & \textbf{33.2}$^{\textcolor{Green}{+3.3}}$
         & \cellcolor{col33}\textbf{65.9}$^{\textcolor{Green}{+6.7}}$ \\
\bottomrule
\end{tabular}}
\end{table}
}

\begin{table}[ht]
\centering
\caption{\small
{\bf POC improves non-VSR tasks.} We test POC on popular few-shot recognition benchmarks using the same expert and LMM as in \cref{tab:abl_POC_v2}. 
POC yields consistent strong accuracy gains over the FSL expert across all benchmarks, highlighting its strong generalization.
}
\vspace{-2mm}
\setlength{\tabcolsep}{0.25em}
\label{tab:nonvsr_generalization}
\scalebox{0.85}{
\begin{tabular}{lllll>{\columncolor{col33}}l}
\toprule
dataset & Aircraft \cite{aircraft} & DTD \cite{dtd} & Food \cite{food} & Cars \cite{cars} & mean acc. \\
\midrule
\rowcolor{gray!15}FSL Expert \cite{liu2025few} & 40.4 & 74.4 & 79.0 & 87.5 & 70.3 \\
POC (ours) & 
\textbf{53.3}$^{\textcolor{Green}{+12.9}}$ &
\textbf{78.6}$^{\textcolor{Green}{+4.1}}$ &
\textbf{86.0}$^{\textcolor{Green}{+7.0}}$ &
\textbf{91.4}$^{\textcolor{Green}{+3.9}}$ &
\textbf{77.3}$^{\textcolor{Green}{+7.0}}$ \\
\bottomrule
\end{tabular}}
\end{table}

{
\setlength{\tabcolsep}{0.7em}
\begin{table}[t]
\centering
\caption{\small
{\bf Comparison of test accuracy between different $k$ values.} 
Using the same expert model and LMM as \cref{tab:abl_POC_v2}, we provide per-dataset POC accuracy with increasing $k$ values.
Results show that increasing $k$ generally improves the test accuracy over the expert model (top-1), while the gains saturate around $k = 10$.
By default, our experiments set $k = 5$.
Note that Mollusca has only 7 classes; hence, its top-10 and top-15 accuracies are the same as top-7.
\textcolor{Green}{Superscripts} denote the accuracy gains over the expert model. 
}
\vspace{-2mm}
\label{tab:ablate_topk_detail}
\scalebox{0.85}{
\begin{tabular}{lccccc>{\columncolor{col33}}l}
\toprule
 & Aves & Inse. & Weeds & Moll. & Fungi & mean acc. \\
\midrule
\rowcolor{gray!15}
top-1 (expert) & 58.2 & 63.8 & 80.7 & 63.6 & 29.9 & \cellcolor{col33}59.2 \\
top-3 & 67.1 & 69.4 & 86.6 & \textbf{69.2} & 31.0 & 64.7$^{\textcolor{Green}{+5.5}}$ \\
top-5  & 69.4 & 70.5 & 87.7 & \textbf{69.2} & 31.1 & 65.6$^{\textcolor{Green}{+6.3}}$ \\
top-7 & 70.9 & 70.5 & \underline{89.1} & \underline{68.5} & \textbf{31.6} & 66.1$^{\textcolor{Green}{+6.9}}$ \\
top-10 & \underline{72.4} & \underline{70.7} & \textbf{89.7} & \underline{68.5} & \underline{31.5} & \textbf{66.5}$^{\textcolor{Green}{+7.3}}$ \\
top-15          & \textbf{73.0} & \textbf{71.5} & 85.3 & \underline{68.5} & 31.3 & \underline{65.9}$^{\textcolor{Green}{+6.7}}$ \\
\bottomrule
\end{tabular}}
\end{table}
}

\textbf{Analysis of hyperparameter $k$ for POC.}
\cref{tab:ablate_topk_detail} presents per-dataset accuracy with different $k$ values. Results show that increasing $k$ generally improves performance, as it is more likely that the ground truths are contained in the top-$k$ predictions.

\textbf{Prompting with taxonomy or text attributes brings limited gains.}
\cref{tab:abl_POC_v2} extends \cref{tab:ablation_mainpaper} in the main paper by 
exploring additional auxiliary information, such as taxonomy and text descriptions of visual attributes, for POC prompting.
Specifically, based on the top-5 class names with confidences scores, further adding taxonomy yields notable improvement. In contrast, adding text attributes provides no benefit and can even degrade performance.
Importantly, adding few-shot images performs best, suggesting that LMMs learn better with visual input than with text input. Finally, when adopting a re-rank strategy, the benefit of taxonomy diminishes.
The results highlight the effectiveness of POC's multimodal prompt.

\textbf{Prompting with visual examples of all classes underperforms POC.} 
For datasets with a small number of classes (e.g., Mollusca and Weeds), one may hypothesize that directly prompting an LMM with both all class names and few-shot visual examples could suffice, without relying on an expert model. 
To investigate this, we construct a multimodal prompt that provides: (i) the test image, (ii) the full list of class names in the dataset, and (iii) few-shot visual examples for each class. \cref{tab:allnames_visuals} reports the results. 
Although adding visual examples improves performance compared to prompting with class names, this strategy still underperforms the FSL expert model. 
In contrast, our POC significantly outperforms all these alternatives. This is likely because the expert model's top-$k$ predictions help narrow down the decision space, emphasizing the necessity of post-hoc processing.

{
\setlength{\tabcolsep}{0.35em}
\begin{table}[t]
\centering
\small
\caption{\small
{\bf Comparison of top-k prompting with various auxiliary contexts.}
We run POC with the FSL expert trained by finetuning the VLM OpenCLIP \cite{cherti2023reproducible} ViT-B/32's visual encoder with 16-shot labeled data for each dataset \cite{liu2025few} and the LMM Qwen-2.5-VL-7B-Instruct \cite{qwen2.5-vl}. We begin by prompting the LMM with the top-5 predicted class names from the expert, further enriched with confidence scores, taxonomy, text descriptions of visual attributes, few-shot visual examples, and a re-ranking strategy.
Results confirm that our POC's multimodal prompt design achieves the best performance.
\textcolor{Green}{Superscripts} denote the accuracy gains over the expert model. 
}
\vspace{-2mm}
\label{tab:abl_POC_v2}
\scalebox{0.85}{
\begin{tabular}{lcccccl}
\toprule
method & Aves & Inse. & Weeds & Moll. & Fungi & \cellcolor{col33}mean acc. \\
\midrule
\rowcolor{gray!15}
FSL Expert \cite{liu2025few} & 58.2 & 63.8 & 80.7 & 63.6 & \underline{29.9} & \cellcolor{col33}59.2  \\
POC w/ top-5 cls names & 63.8 &40.1  &81.3  &46.9  &16.2  &\cellcolor{col33}49.6$^{\textcolor{Red}{-9.6}}$ \\

\quad + few-shot images &68.9 &61.1 &88.6 &56.4 &27.7 &\cellcolor{col33}60.5$^{\textcolor{Green}{+1.3}}$ \\

\quad + confidences & 68.8 & 59.6 & 85.8 & 58.7 & 25.1 & \cellcolor{col33}59.6$^{\textcolor{Green}{+0.4}}$ \\

\quad \quad + taxonomy & 68.3 & 66.5 & 86.9 & 63.2 & 24.5 & \cellcolor{col33}61.9$^{\textcolor{Green}{+2.7}}$ \\

\quad \quad + text attributes & 68.3 & 44.6 & 80.8 & 56.5 & 16.3 & \cellcolor{col33}53.3$^{\textcolor{Red}{-5.9}}$ \\

\quad \quad + few-shot images & \textbf{70.2} & 67.4 & \underline{88.5} & 63.4 & 29.5 & \cellcolor{col33}63.8$^{\textcolor{Green}{+4.6}}$ \\

\quad \quad \quad + taxonomy &69.0 &69.7 &\textbf{88.6} &67.6 &29.8 & \cellcolor{col33}64.9$^{\textcolor{Green}{+5.7}}$ \\

\quad \quad \quad \textbf{+ re-rank (POC)} & \underline{69.4} & \textbf{70.5} & 87.7 & \textbf{69.2} & \textbf{31.1} & \cellcolor{col33}\textbf{65.6$^{\textcolor{Green}{+6.4}}$} \\

\quad \quad \quad + taxonomy + re-rank &\underline{69.4} &\underline{70.3} &88.1 &\underline{68.6} &29.7  & \cellcolor{col33}\underline{65.2}$^{\textcolor{Green}{+6.0}}$ \\

\bottomrule
\end{tabular}}

\vspace{-2mm}
\end{table}}

{
\setlength{\tabcolsep}{0.6em}  
\begin{table}[t]
\centering
\small
\caption{
\small
\textbf{Performance of prompting LMM with all class names and visual examples.} As before, we train the FSL expert model by finetuning the VLM OpenCLIP \cite{cherti2023reproducible} ViT-B/32 model's visual encoder on 16-shot labeled data from each dataset. We then run POC with the LMM Qwen-2.5-VL-7B (Qwen). Superscripts denote gains (\textcolor{Green}{green}) or degradations (\textcolor{Red}{red}) relative to the FSL expert.
}
\vspace{-2mm}
\label{tab:allnames_visuals}
\scalebox{0.85}{
\begin{tabular}{lll}
\toprule
&Mollusca (7 classes) & Weeds (20 classes) \\
\midrule
\rowcolor{gray!15}
FSL Expert \cite{liu2025few} &63.6 &80.7 \\
LMM (w/ all names) &45.2$^{\textcolor{Red}{-18.4}}$  &78.1$^{\textcolor{Red}{-2.6}}$  \\
LMM (w/ all names + visuals) &56.5$^{\textcolor{Red}{-7.1}}$  &80.0$^{\textcolor{Red}{-0.7}}$\\
POC (ours) & \textbf{69.2$^{\textcolor{Green}{+5.6}}$} &\textbf{87.7$^{\textcolor{Green}{+7.0}}$} \\
\bottomrule
\end{tabular}}
\end{table}}

\begin{figure*}[t]
    \centering
    \includegraphics[width=0.9\linewidth, clip=true, trim=0mm 0mm 0mm 0mm]{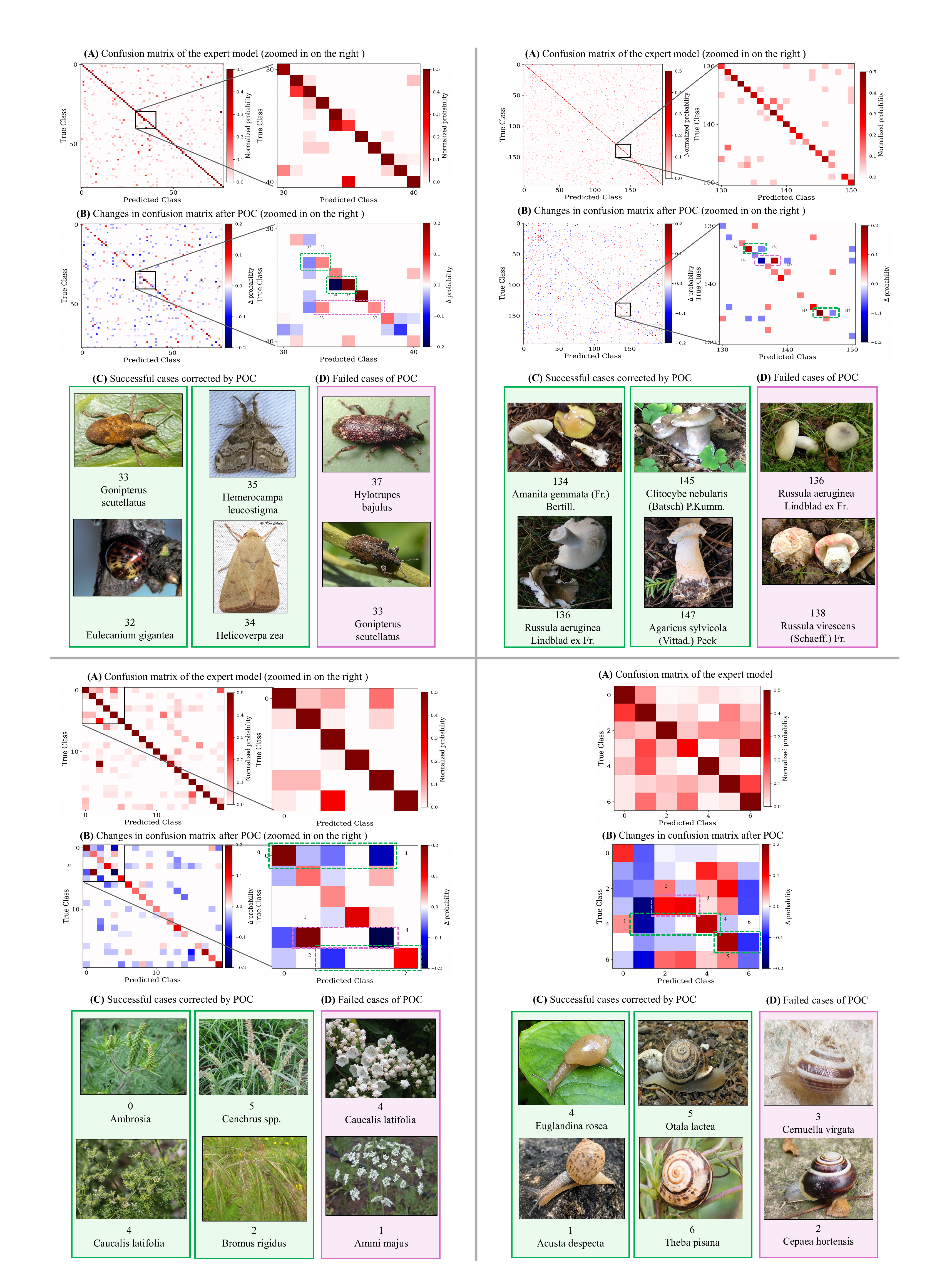}
    
\vspace{-2mm}
    \caption{\small
    \textbf{Visualization of the confusion matrix on more VSR benchmarks}, including Insecta (top-left), Fungi (top-right), Weeds (bottom-left), Mollusca (bottom-right).
    For each benchmark, we show the (A) confusion matrix of the FSL expert model, (B) changes in the confusion matrix after POC, (C) and (D) successful and failed cases by POC.
    Results show that POC can leverage LMM to correct visually confusing species that the FSL expert model fails to distinguish, while it may also fail in some other cases.
    }
    \label{fig:confusion_matrix_combine}
\end{figure*}

\textbf{More visualizations of the confusion matrix.}
The following \cref{fig:confusion_matrix_combine} supplements the main paper with more visual examples from other benchmarks. 
The results highlight that LMMs can help distinguish some confusing species pairs that the expert model struggles with, thereby improving the accuracy.

\end{document}